\documentclass{article}

\usepackage[eandd,final]{neurips_2026}

\usepackage[utf8]{inputenc} % allow utf-8 input
\usepackage[T1]{fontenc}    % use 8-bit T1 fonts
\usepackage{hyperref}       % hyperlinks
\usepackage{url}            % simple URL typesetting
\usepackage{booktabs}       % professional-quality tables
\usepackage{amsfonts}       % blackboard math symbols
\usepackage{nicefrac}       % compact symbols for 1/2, etc.
\usepackage{microtype}      % microtypography
\usepackage{xcolor}         % colors
\usepackage{makecell}
\usepackage{pgfplots}
\pgfplotsset{compat=1.18}
\usepackage{wrapfig}
\usepackage{subcaption}
\usepackage{graphicx}

\usepackage{natbib}

\hypersetup{
colorlinks=true
}

\usepackage{amsmath, amssymb, amsthm}
\usepackage{graphicx}
\usepackage{booktabs}
\usepackage{multirow}
\usepackage{rotating}
\usepackage{xcolor}
\usepackage{enumitem}
\usepackage{hyperref}
\usepackage{tcolorbox}

\usepackage{tikz}
\usetikzlibrary{positioning, calc, shapes, shapes.geometric, shadows.blur, arrows.meta, decorations.pathreplacing, patterns}

\usepackage{pifont}
\newcommand{\cmark}{\textcolor{green!55!black}{\ding{51}}}
\newcommand{\xmark}{\textcolor{red!70!black}{\ding{55}}}

\newtheorem{definition}{Definition}

\newcommand{\Accq}{\mathrm{Acc}_{\mathrm{q}}}
\newcommand{\Accg}{\mathrm{Acc}_{\mathrm{g}}}

\newcommand{\triad}{\textsc{Triad}}

\title{TRIAD: Benchmarking Omni-Modal Ambiguity\\ in Multimodal Large Language Models
}

\author{%
  \textbf{Zhaolu Kang}$^{1,2,*}$, \textbf{Yidi Wang}$^{2,*}$, \\\textbf{Yile Li}, \textbf{Siqi Zeng},
  \textbf{Yingjie He},
  \textbf{Richeng Xuan}$^{1,\dagger}$ ,\textbf{Zhichao Hu}$^{1,\dagger}$ \\
  $^{1}$Hunyuan, Tencent \quad $^{2}$Peking University  \\
  \texttt{zlkang25@stu.pku.edu.cn}
}

\begin{document}

\maketitle

\renewcommand{\thefootnote}{\fnsymbol{footnote}}
\footnotetext[1]{Equal Contribution. $^{\dagger}$Corresponding Author.}
\renewcommand{\thefootnote}{\arabic{footnote}}

\begin{abstract}
Ambiguity is a central feature of natural communication: a sentence, image, or sound may remain underspecified until evidence from the other channels is considered. Yet current multimodal benchmarks mostly test perception, recognition, or reasoning over already determinate inputs, leaving unclear whether omni-modal large language models can resolve ambiguity distributed across text, vision, and audio. We introduce \triad{}, a benchmark for omni-modal ambiguity resolution. Each item is a text--image--audio triplet paired with a question and a set of answer options, and is constructed so that the full triplet determines a unique answer while removing any one modality makes the set of consistent options non-unique. \triad{} contains $327$ hand-written queries in $86$ scene groups and covers $18$ ambiguity categories across the three modalities. Evaluating $14$ omni-modal systems reveals that strong item-level performance does not translate into robust scene-level disambiguation: the best model still trails humans by over $20$ points in group accuracy. Removing answer options widens the gap further, driving model group accuracy to the single digits while humans remain robust. Leave-one-out and transcript/description counterfactuals further show that current systems often underuse audio cues when text and image are present. \triad{} reframes omni-modal evaluation around disambiguation rather than recognition, and provides a taxonomy, benchmark, and diagnostic protocol for measuring this capability.

\end{abstract}
% --- Main body ---

\section{Introduction}
\label{sec:intro}

\begin{figure}[t]
  \centering
  \includegraphics[width=\textwidth]{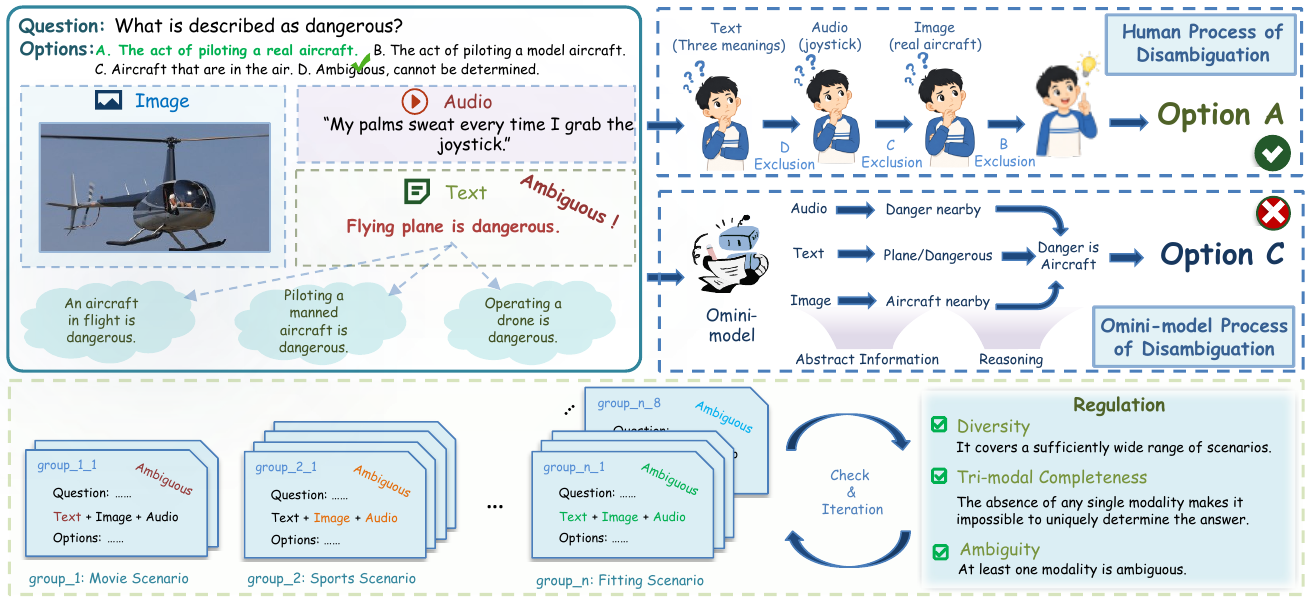}
  \vspace{-1.8em}
  \caption{Overview of \triad{}. Each query is constructed so that text, image, and audio mutually disambiguate one another: only the full triplet determines a unique answer, while any two-modality view remains ambiguous. The benchmark contrasts human-like cross-modal disambiguation with model behavior under partial modality cues, and organizes examples into checked scene groups.}
  \label{fig:triadhead}\textit{}
    \vspace{-1.2em}
\end{figure}

Interpreting multimodal communication often requires more than recognizing what is present in each channel. A sentence may leave several readings open, an image may constrain but not determine the scene, and audio may provide prosody, speaker information, or environmental context that selects the intended interpretation. For omni-modal large language models, this raises a question that is distinct from standard perception or question answering: can a model resolve an ambiguity that remains underdetermined unless text, image, and audio are considered together?

Existing benchmarks do not isolate this capability. Ambiguity-focused datasets mainly study visual ambiguity or text--image disambiguation~\citep{guan2024hallusionbench,luo2024codis,wang2025mucar,ma20243am,nam2025vaguevisualcontextsclarify}, while audio-visual benchmarks usually evaluate recognition of events, objects, speech, or sounds. Broad omni-modal evaluations cover more tasks and modalities~\citep{fu2025mmecomprehensiveevaluationbenchmark,liu2024mmbenchmultimodalmodelallaround,li2025omnibenchfutureuniversalomnilanguage,yang2024airbenchbenchmarkinglargeaudiolanguage,wang2025omnimmicomprehensivemultimodalinteraction}, but they rarely make ambiguity a tri-modal mutual-disambiguation problem: text, image, and audio may each narrow the interpretation, yet the intended answer becomes unique only when all three are present.

We introduce \triad{}, a benchmark for tri-modal mutual disambiguation. As illustrated in Figure~\ref{fig:triadhead}, each item contains a text--image--audio triplet, a query, and answer options. The complete triplet has one correct answer, but every two-modality view leaves multiple options consistent with the remaining evidence. Related queries are further organized into scene groups that share a core context and perturb one modality, so evaluation can test whether models track the cue that changes rather than reuse a fixed scene prior.

\triad{} has three defining properties: a unified $18$-category taxonomy over linguistic, visual, and acoustic ambiguity; an item-level mutual-disambiguation check ensuring that only the full triplet resolves the ambiguity; and scene groups whose sister queries perturb one modality at a time. These choices make the benchmark diagnostic of cross-modal disambiguation rather than ordinary recognition or question answering. \triad{} contains $327$ queries in $86$ scene groups across English and Chinese settings. Detailed tier distributions, media-source details, and admission procedures are reported in Sections~\ref{sec:benchmark} and~\ref{sec:construction}.

We evaluate $14$ omni-modal systems under three prompt formats and eight modality conditions. Two results define the current gap. First, the best model reaches $71.2$ group-wise accuracy under multiple choice, while the human baseline is $91.6$; the same model reaches $90.4$ query accuracy, showing that local success on individual sisters does not imply consistency at the scene level. Second, removing the option list causes every model to fall to the low single digits in group-wise accuracy, while humans remain substantially more robust. The remaining experiments locate where this gap appears across modalities, ambiguity tiers, prompt formats, and error types.

The paper makes three contributions:
\begin{itemize}[leftmargin=*, noitemsep, topsep=0pt]
  \item A taxonomy of multimodal ambiguity with $18$ categories across text, image, and audio, including audio categories for prosody, speaker traits, phonetic form, environmental sound, and non-speech symbols.
  \item \triad{}, a benchmark of $327$ queries in $86$ scene groups, with a tri-modal mutual-disambiguation property checked by five annotators and a reference checker released for future extensions.
  \item An evaluation of $14$ omni-modal systems under three prompt formats and eight modality conditions, with human and random baselines, category and tier breakdowns, and counterfactual inputs that replace audio with a transcript and image with a description.
\end{itemize}

\section{Related work}
\label{sec:related}

Multimodal ambiguity has been studied in settings where one modality exposes or resolves uncertainty in another. HallusionBench~\citep{guan2024hallusionbench} probe visual illusions and counterintuitive images; CODIS~\citep{luo2024codis} and MUCAR~\citep{wang2025mucar} pair ambiguous visual content with textual contexts; 3AM~\citep{ma20243am} and UNPIE~\citep{Scaccia2025} examine ambiguous instructions or lexical meanings under visual grounding. These benchmarks make ambiguity an explicit evaluation target, but the ambiguity is primarily visual, textual, or text--image. They therefore differ from the tri-modal setting studied here, where text, image, and audio all carry evidence and no two-modality subset is sufficient to determine the intended answer. Appendix~\ref{app:bench-comparison} provides a detailed comparison along these axes.

Audio-visual and omni-modal benchmarks broaden the input space but usually evaluate recognition, transcription, or general question answering rather than mutual disambiguation. AVQA~\citep{fu2025mmecomprehensiveevaluationbenchmark} and AVHBench~\citep{sungbin2025avhbenchcrossmodalhallucinationbenchmark} ask models to identify events, objects, speakers, instruments, or hallucinated audio-visual content. Recent omni-modal systems, including GPT-4o~\citep{openai2024gpt4ocard}, Gemini~2~\citep{geminiteam2025geminifamilyhighlycapable}, Qwen2.5-Omni~\citep{xu2025qwen25omnitechnicalreport}, MiniCPM-o and Unified-IO-2~\citep{lu2023unifiedio2scalingautoregressive}, motivate broad evaluations such as OmniBench~\citep{li2025omnibenchfutureuniversalomnilanguage}, AIR-Bench, and OmniMMI~\citep{wang2025omnimmicomprehensivemultimodalinteraction}. \triad{} targets a narrower failure mode: whether an omni-modal model can combine three incomplete cues into one unique interpretation, rather than answer from a single dominant modality.

Speech and linguistics research further shows that audio can change meaning in ways that are not recoverable from text alone. Prosody, speaker information, phonetic form, environmental sound, and non-speech symbols can signal intent, affect, reference, or scene context~\citep{HIRSCHBERG200231}. In \triad{}, these audio cues are not treated as standalone labels; they serve as one side of a mutual-disambiguation problem together with language and vision. The resulting taxonomy organizes ambiguity categories across the three modalities while keeping the evaluation focused on whether the full triplet is necessary for resolution.

\section{The \triad{} benchmark}
\label{sec:benchmark}

\triad{} is designed to test a specific capability: resolving an interpretation that remains ambiguous unless text, image, and audio are used together. This section defines the task, the mutual-disambiguation criterion, the taxonomy used for diagnosis, and the group-wise evaluation protocol.

\subsection{Task formulation}
\label{subsec:task-formulation}

Each query is a tuple $x = (\mathcal{C}, \mathcal{I}, \mathcal{A}, \mathcal{Q}, \mathcal{Y})$, where $\mathcal{C}$ is a textual context, $\mathcal{I}$ is an image, $\mathcal{A}$ is an audio clip, $\mathcal{Q}$ is a question, and $\mathcal{Y}$ is a finite option set with $|\mathcal{Y}| \ge 2$. The model must select the intended answer $y^{*} \in \mathcal{Y}$.

The benchmark is not meant to reward recognition from a single dominant channel. Instead, each item is constructed so that every modality narrows the set of plausible answers, while none of the two-modality views is sufficient. Text may leave a reference open, the image may constrain the scene without fixing the intended interpretation, and audio may supply prosody, speaker cues, phonetic evidence, or environmental context that changes the reading. The answer becomes unique only after the three sources are combined.

\subsection{Mutual-disambiguation criterion}
\label{sec:benchmark-wellposedness}

Let $\mathcal{M}=\{\mathcal{T},\mathcal{V},\mathcal{A}\}$ denote the three modalities. For a subset $S \subseteq \mathcal{M}$, let $x_S$ be the input restricted to the modalities in $S$, and let
\begin{align}
\mathcal{Y}_S(x)=\{\,y\in\mathcal{Y}: y \text{ is consistent with } x_S\,\}.
\end{align}
\begin{definition}[Well-posed tri-modal disambiguation]
\label{def:wellposed}
A query $x$ is well posed if the full triplet determines exactly one answer, while every strict two-modality view remains non-unique:
\begin{align}
|\mathcal{Y}_{\mathcal{M}}(x)| &= 1, \\
\forall S \subset \mathcal{M},\ |S|=2: \quad |\mathcal{Y}_S(x)| &> 1.
\end{align}
\end{definition}

The first condition rules out underdetermined examples: the full input must support one labelled answer. The second condition rules out reducible examples: if text--image, text--audio, or image--audio already identifies the answer, then the item does not test tri-modal disambiguation. This criterion is about the option set, not about whether each modality is visibly ambiguous in isolation. A modality can be locally clear and still be necessary because it eliminates an option that the other two modalities leave open.
% For example, an item asking who borrowed an umbrella is invalid if text, image, and audio together still leave two people consistent with the answer. It is also invalid if text and image already settle the answer while the audio only decorates the scene. 
Released items are checked with a five-annotator subset checklist described in Section~\ref{subsec:verifier}; Appendix~\ref{app:irreducibility} gives the reference checker used for future extensions.

\subsection{Ambiguity taxonomy}
\label{subsec:taxonomy}

\triad{} uses an $18$-category taxonomy to describe where ambiguity enters the item and to support error analysis after evaluation. The taxonomy has six leaf categories for each modality. Text labels cover anaphoric, syntactic, logical, pragmatic, temporal, and lexical ambiguity. Image labels cover semiotic, grounding, spatial, kinematic, observational, and contextual ambiguity. Audio labels cover referential, prosodic, indexical, phonetic, environmental, and symbolic ambiguity.

Each item receives one text label, one image label, and one audio label. Consequently, each modality-specific leaf distribution contains $327$ labels, one per query. These labels do not imply that the three modalities are independently ambiguous to the same degree; they identify the kind of evidence each channel contributes to the mutual-disambiguation problem. The full definitions and examples are provided in Appendix~\ref{app:taxonomy}.

\subsection{Scene groups and metrics}
\label{subsec:groups-metrics}

A scene group $\mathcal{G}$ contains $k \in \{2,3,4,8\}$ sister queries. Sisters share a core scenario and answer space, but perturb one modality so that the correct answer changes with the altered cue; the question template and option semantics are kept fixed whenever possible, with minimal adjustments when needed to keep all sisters well posed. In image-perturbed groups, images vary while text and audio stay fixed. In audio-perturbed groups, audio clips vary while text and image stay fixed. Mixed groups perturb different modalities across sisters.

This format discourages models from answering from a static scene prior. A model must track which modality changed and update the answer consistently across the group. We therefore report both query accuracy and group accuracy. For predictions $\hat{y}(x)$ over $n_q=327$ queries and $n_g=86$ groups,
\begin{align}
\Accq &= \frac{1}{n_q}\sum_x \mathbb{1}\!\left[\hat{y}(x)=y^{*}(x)\right], \\
\Accg &= \frac{1}{n_g}\sum_{\mathcal{G}} \mathbb{1}\!\left[\forall x\in\mathcal{G}:\hat{y}(x)=y^{*}(x)\right].
\end{align}
$\Accq$ measures item-level success, while $\Accg$ counts a group as correct only when every sister query is answered correctly. We use $\Accg$ as the primary metric because it captures whether a model follows the perturbation across the scene.

\subsection{Dataset overview}
\label{subsec:stats}

\begin{wrapfigure}{r}{0.48\textwidth}
  \vspace{-1.2em}
  \centering
  \includegraphics[width=0.47\textwidth]{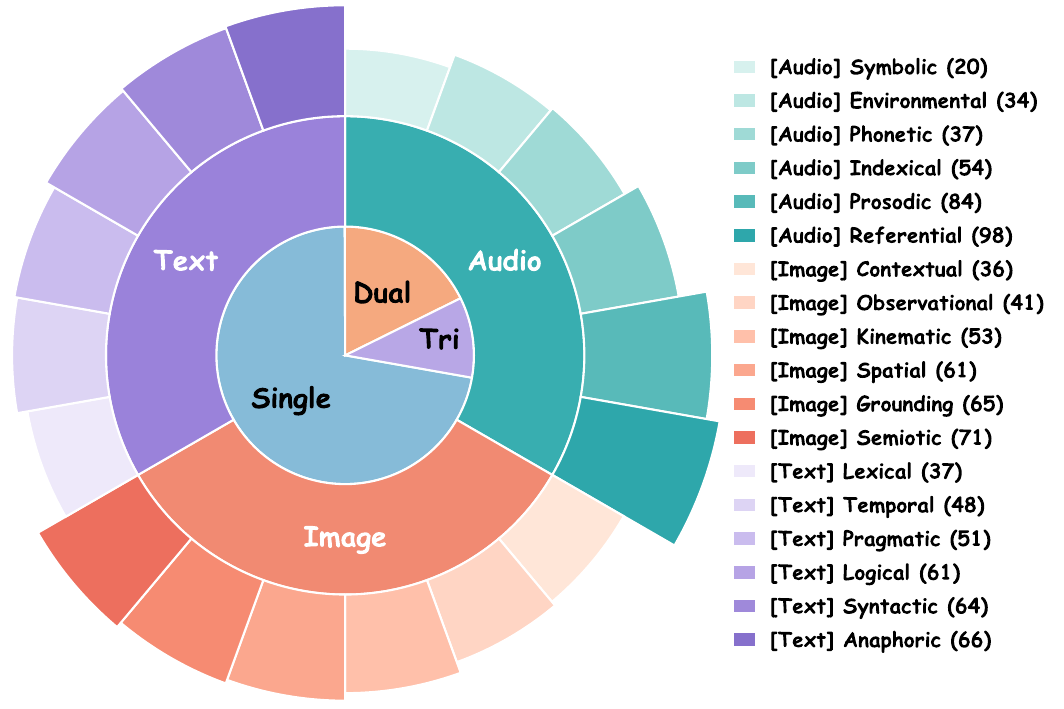}
  \caption{\textbf{TRIAD distribution overview.} The inner disc shows ambiguity tiers, and the outer rings show the leaf-category distribution for text, image, and audio.}
  \label{fig:donut}
  \vspace{-1.4em}
\end{wrapfigure}

The released benchmark contains $327$ queries in $86$ scene groups. Group sizes are $\{2,3,4,8\}$, giving a mean of $3.8$ queries per group. The data covers English and Chinese settings, with $184$ English queries and $143$ Chinese queries. Option sets contain $2$, $3$, $4$, $5$, or $8$ choices.

Figure~\ref{fig:donut} summarizes the benchmark distribution in a compact form. The inner disc shows the ambiguity-tier axis: $230$ items contain one internally ambiguous modality, $59$ contain two, and $38$ contain three. This tier is descriptive rather than a relaxation of the task: every item must satisfy Definition~\ref{def:wellposed}. The outer rings show the taxonomy axis, with one leaf label per modality for every query; consequently, each modality-specific ring sums to $327$ labels. Detailed leaf definitions and counts are reported in Appendix~\ref{app:taxonomy}.

\subsection{Intended uses and non-uses}
\label{subsec:intended-uses}

\triad{} is a diagnostic benchmark, not a training set. Its tags for modality, category, tier, and scene group, together with the language split, are intended to support controlled error analysis. We recommend reporting $\Accg$ as the primary metric, $\Accq$ as a secondary metric, and open-ended scores as a lower bound on free-response disambiguation (Section~\ref{subsec:closed-vs-open}). \triad{} is not a general recognition benchmark, not a transcription benchmark, and not a replacement for broad multimodal evaluation; it isolates whether a model can combine three individually insufficient cues into one unique interpretation.

\section{Construction pipeline}
\label{sec:construction}

Four exclusion rules apply throughout construction: no item may be answerable from a smaller modality subset, English text must be grammatical, any AI generated image must pass manual artefact screening, and no content may involve violence, privacy violations, or other ethical risks.

\subsection{Seeding strategy}

Scene groups are constructed through three complementary approaches to ensure diverse and challenging ambiguity patterns. For text-based construction, annotators create sentences and questions that inherently contain linguistic ambiguities, then extend them with audio cues that alter the interpretation. For image-based construction, annotators select or create images with visual ambiguities and design questions that require integrating textual and prosodic information. For audio-based construction, annotators begin with naturally ambiguous audio clips and build appropriate textual and visual contexts around them.

\subsection{Per-modality content acquisition}

Sentences are written by annotators and checked by two reviewers for grammaticality and idiom. Within a group, sisters share a core scenario and answer space; question templates and option semantics are kept fixed when possible, with only minimal wording or cardinality adjustments when needed to keep each sister well posed.

Images are predominantly real photographs sourced from public repositories with research-friendly licenses. A small fraction uses AI generated images, only when no real photograph captures the intended ambiguity. Two annotators screen each such image for common artefacts such as anatomical errors, malformed text, or inconsistent surfaces, and source details are documented in the release material so users can filter them when available.

Audio clips are required to sound natural. Sources are prioritized in the following order: natural recordings from permissively licensed corpora, human voice acting, and acoustic editing of natural recordings. Text-to-speech audio is used only when a human edits the prosody and confirms that the result sounds natural. Audio source details are documented with the dataset.

\subsection{Well-posedness review}
\label{subsec:verifier}

Well-posedness (Definition~\ref{def:wellposed}) is checked by five annotators rather than by an automated gate. Each candidate is reviewed with a modality subset checklist. For every strict subset $S \subsetneq \mathcal{M}$ with $|S| \in \{1,2\}$, reviewers record at least two options that remain consistent with $x_S$. For the full triplet, exactly one option must remain. The checklist targets the two failure modes in Section~\ref{sec:benchmark-wellposedness}: items with no unique answer, and items in which one modality is decorative.

Across authoring, exclusion checks, the subset checklist, independent answer checking, and group integration (Appendix~\ref{app:annotator}), approximately $40$ percent of candidate items are admitted. A Python reference checker for well-posedness verification is provided in the supplementary material for reproducibility.

\subsection{Annotator pool and inter-annotator agreement}

Five annotators participated: three with NLP backgrounds and two with linguistics training. Each query was written by one annotator and reviewed by the other four. A submission was admitted only if the labelled answer was correct, the item was distinct from existing groups, all exclusion rules passed, and the well-posedness checklist in Section~\ref{subsec:verifier} was satisfied. On a held-out audit subset, agreement on the labelled answer is Cohen's $\kappa = 0.84$, and agreement on modality ambiguity tags is $\kappa = 0.79$. Appendix~\ref{app:annotator} gives the instructions, checklist, and agreement by modality and category.

\subsection{Case examples}
\label{subsec:case-examples}

\begin{figure}[t]
\centering
\vspace{-2em}
\includegraphics[width=0.9\textwidth]{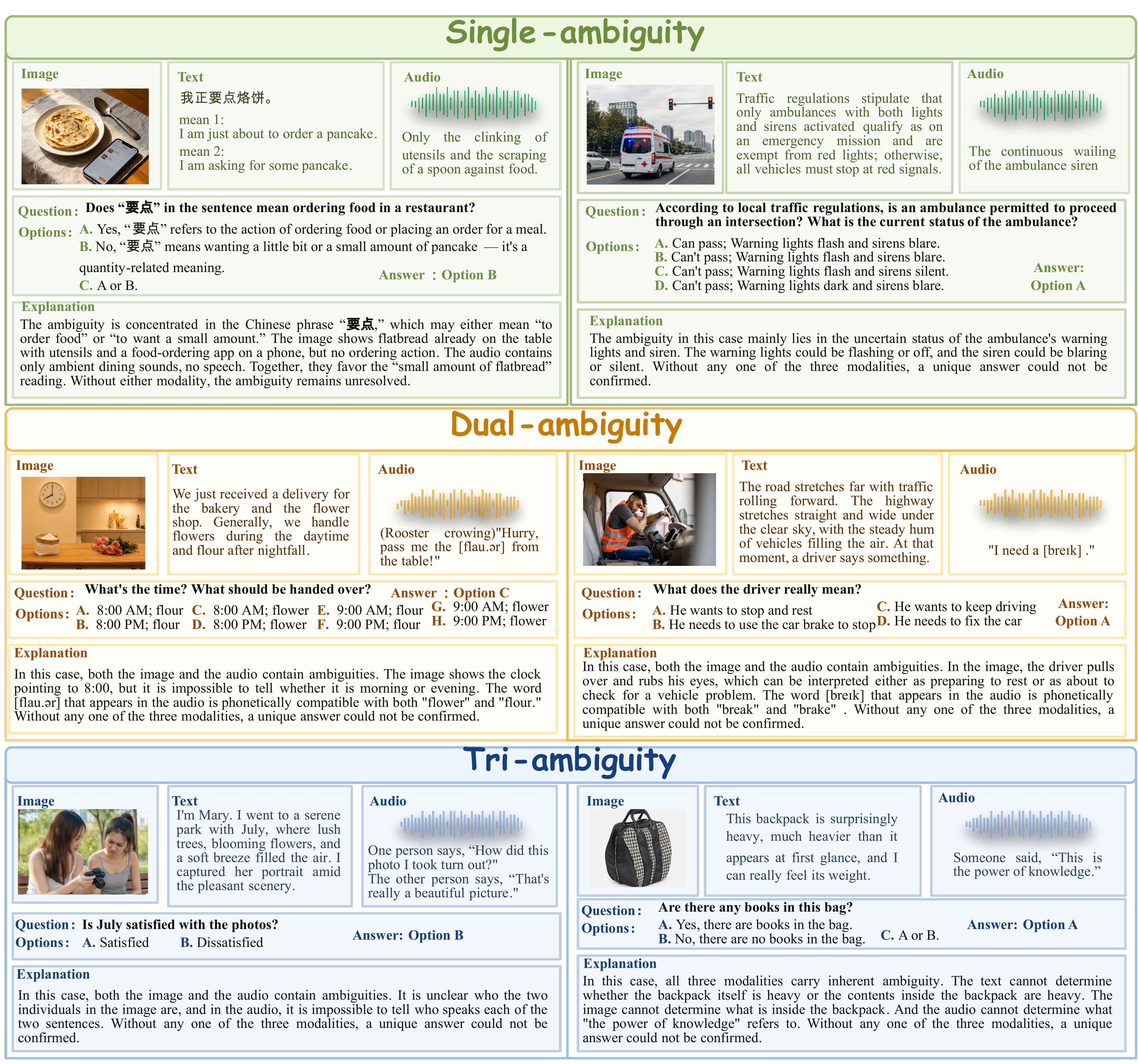}
\caption{\textbf{Case examples across ambiguity tiers.} The dataset contains items with varying levels of internal ambiguity across modalities. Each case shows how different modality combinations create distinct disambiguation challenges.}
\vspace{-1.8em}
\label{fig:case-examples}
\end{figure}

Figure~\ref{fig:case-examples} shows six representative cases from the dataset, illustrating the three ambiguity tiers: single-ambiguity (one modality ambiguous), multi-ambiguity (two modalities ambiguous), and tri-ambiguity (all three modalities ambiguous). These examples demonstrate how the well-posedness property ensures that only the full triplet uniquely determines the answer.

\section{Experiments and Analysis}
\label{sec:exp}

The experiments test three questions: whether models resolve tri-modal
ambiguity, whether they stay consistent across sister queries, and whether their
answers are controlled by the intended modality evidence rather than by option
priors.

\subsection{Setup}

\paragraph{Models and prompts.}
We evaluate six proprietary and eight open-source omni-modal systems that accept
text, image, and audio directly. 
The proprietary models are Gemini~2.5 Pro, Gemini~2.5 Flash, Gemini~3 Pro, Gemini~3 Flash, Gemini~3.1 Pro~\citep{geminiteam2025geminifamilyhighlycapable}, and GPT-4o~\citep{openai2024gpt4ocard}. The open source models are Qwen3.5-Omni-Plus, Qwen3-Omni-30B, Qwen2.5-Omni-7B, Qwen2.5-Omni-3B~\citep{xu2025qwen25omnitechnicalreport}, MiniCPM-o-4.5, OmniVinci-9B, Ola-7B, and Baichuan-Omni-1.5. 
The main comparison uses native tri-modal input
rather than bi-modal fallbacks. Each model is evaluated with three prompt
formats: Standard multiple choice, Chain-of-thought multiple choice, and
Open-ended answering without the option list. All runs use temperature $0$ and
one sample per query.

\paragraph{Input conditions.}
The main condition provides the full text--image--audio triplet. We also evaluate
empty-input, single-modality, and leave-one-out views. Because \triad{} admits an
item only when no two modalities determine the answer, high ablated scores should
be read as evidence of priors or shortcut use, not evidence that the omitted
modality is unnecessary.

\paragraph{Metrics and grading.}
We report query accuracy $\Accq$ and group accuracy $\Accg$. $\Accq$ measures
isolated answers, while $\Accg$ requires every sister query in a scene group to
be answered correctly and is therefore our primary metric. Standard and
Chain-of-thought outputs are graded by exact option match. Open-ended responses
are graded by two independent judge models, and we report the mean judge score;
empty, malformed, or refusal outputs are counted as incorrect in the
full-denominator open-ended scores. Human scores are measured on a
cost-controlled $100$-query stratified subset, using the same subset for
Standard multiple choice and Open-ended answering. Model and random results are
computed on the full $327$-query benchmark unless otherwise stated. The random
baseline uses empirical option counts and group sizes.

\subsection{Main diagnostic results}

\begin{table}[t]
\centering
\setlength{\tabcolsep}{1.6pt}
\renewcommand{\arraystretch}{0.96}
\scriptsize
\resizebox{\textwidth}{!}{%
\begin{tabular}{l cc cc cc | cc cc cc cc cc cc cc}
\toprule
 & \multicolumn{6}{c|}{\textbf{Full input}}
 & \multicolumn{14}{c}{\textbf{Group-accuracy controls}} \\
\cmidrule(lr){2-7}\cmidrule(lr){8-21}
 & \multicolumn{2}{c}{Std}
 & \multicolumn{2}{c}{CoT}
 & \multicolumn{2}{c|}{Open}
 & \multicolumn{2}{c}{None}
 & \multicolumn{2}{c}{Text only}
 & \multicolumn{2}{c}{Image only}
 & \multicolumn{2}{c}{Audio only}
 & \multicolumn{2}{c}{No-Text}
 & \multicolumn{2}{c}{No-Image}
 & \multicolumn{2}{c}{No-Audio} \\
\cmidrule(lr){2-3}\cmidrule(lr){4-5}\cmidrule(lr){6-7}
\cmidrule(lr){8-9}\cmidrule(lr){10-11}\cmidrule(lr){12-13}\cmidrule(lr){14-15}
\cmidrule(lr){16-17}\cmidrule(lr){18-19}\cmidrule(lr){20-21}
Model
 & q & g & q & g & q & g
 & Std & Open & Std & Open & Std & Open & Std & Open & Std & Open & Std & Open & Std & Open \\
\midrule
Random
 & 24.3 &  2.2 & --   & --   & --   & --
 &  2.2 & -- &  2.2 & -- &  2.2 & -- &  2.2 & -- &  2.2 & -- &  2.2 & -- &  2.2 & -- \\
Human
 & 95.1 & 91.6 & --   & --   & 88.4 & 79.2
 &  2.4 & 0.6 & 18.2 & 6.3 & 15.6 & 5.0 & 11.8 & 4.2 & 22.9 & 8.4 & 70.5 & 44.9 & 60.7 & 36.1 \\
\midrule
\multicolumn{21}{l}{\emph{Proprietary}} \\
Gemini 3.1 Pro
 & \textbf{90.4} & \textbf{71.2} & \textbf{91.1} & \textbf{72.5} & \textbf{18.4} & \textbf{4.6}
 &  0.8 & 0.2 & \textbf{42.6} & \textbf{8.1} & \textbf{8.9} & \textbf{1.3} & \textbf{3.7} & \textbf{0.6} & \textbf{16.4} & \textbf{2.4} & \textbf{57.2} & \textbf{12.4} & \textbf{51.8} & \textbf{10.7} \\
Gemini 3 Pro
 & 89.2 & 68.9 & 90.0 & 70.4 & 16.1 & 3.1
 &  0.7 & 0.1 & 40.1 & 6.4 &  8.2 & 0.9 &  3.4 & 0.5 & 15.6 & 1.8 & 54.7 & 10.2 & 49.4 &  8.5 \\
Gemini 3 Flash
 & 82.5 & 59.4 & 82.0 & 58.1 & 17.6 & 3.5
 &  0.6 & 0.1 & 38.9 & 6.9 &  7.1 & 0.8 &  2.9 & 0.4 & 11.3 & 1.6 & 52.3 & 10.6 & 45.8 &  8.7 \\
Gemini 2.5 Pro
 & 80.7 & 55.3 & 82.4 & 58.0 & 16.9 & 3.3
 &  0.6 & 0.1 & 33.5 & 5.7 &  7.8 & 0.9 &  3.1 & 0.4 & 13.0 & 1.7 & 45.9 &  9.1 & 40.2 &  7.9 \\
GPT-4o
 & 76.3 & 49.8 & 78.1 & 52.9 & 15.4 & 2.8
 &  0.5 & 0.1 & 28.9 & 4.6 &  7.4 & 0.7 &  2.8 & 0.3 & 12.1 & 1.4 & 41.8 &  8.0 & 36.5 &  7.1 \\
Gemini 2.5 Flash
 & 73.2 & 45.6 & 74.1 & 46.8 & 15.8 & 2.9
 &  0.3 & 0.1 & 30.3 & 5.2 &  5.4 & 0.6 &  2.1 & 0.2 &  9.2 & 1.3 & 40.5 &  7.6 & 34.1 &  6.8 \\
\midrule
\multicolumn{21}{l}{\emph{Open-source omni}} \\
Qwen3.5-Omni-Plus
 & 78.0 & 52.8 & 78.2 & 51.4 & 17.9 & 3.1
 &  0.4 & 0.1 & 32.1 & 5.3 &  6.9 & 0.7 &  2.4 & 0.3 & 11.8 & 1.4 & 43.6 &  8.6 & 39.7 &  7.5 \\
Qwen3-Omni-30B
 & 72.5 & 41.7 & 72.8 & 40.9 & 14.7 & 2.3
 &  0.4 & 0.0 & 25.0 & 3.9 &  5.8 & 0.5 &  2.0 & 0.2 & 10.3 & 1.1 & 35.2 &  6.3 & 30.2 &  5.4 \\
MiniCPM-o-4.5
 & 66.2 & 36.4 & 68.0 & 38.2 & 12.3 & 1.6
 &  0.3 & 0.0 & 21.3 & 2.5 &  4.7 & 0.3 &  1.8 & 0.1 &  8.1 & 0.8 & 30.1 &  4.4 & 25.9 &  3.7 \\
Qwen2.5-Omni-7B
 & 61.8 & 30.2 & 63.9 & 32.5 & 11.0 & 1.2
 &  0.3 & 0.0 & 17.2 & 1.8 &  3.9 & 0.2 &  1.5 & 0.1 &  6.5 & 0.5 & 24.6 &  3.1 & 20.8 &  2.6 \\
OmniVinci-9B
 & 58.7 & 26.9 & 58.2 & 25.4 &  9.8 & 1.0
 &  0.3 & 0.0 & 15.0 & 1.4 &  3.5 & 0.2 &  1.3 & 0.1 &  5.7 & 0.4 & 21.7 &  2.6 & 18.4 &  2.2 \\
Ola-7B
 & 55.4 & 23.8 & 56.8 & 24.7 &  9.1 & 0.9
 &  0.2 & 0.0 & 13.2 & 1.1 &  3.0 & 0.1 &  1.1 & 0.0 &  4.9 & 0.3 & 18.9 &  2.1 & 16.1 &  1.8 \\
Qwen2.5-Omni-3B
 & 54.1 & 22.1 & 53.5 & 20.6 &  8.4 & 0.7
 &  0.2 & 0.0 & 11.9 & 0.9 &  2.7 & 0.1 &  1.0 & 0.0 &  4.4 & 0.2 & 17.3 &  1.7 & 14.8 &  1.5 \\
Baichuan-Omni-1.5
 & 52.9 & 20.4 & 52.4 & 19.2 &  7.6 & 0.6
 &  0.2 & 0.0 & 10.6 & 0.7 &  2.4 & 0.1 &  0.9 & 0.0 &  4.0 & 0.2 & 15.8 &  1.5 & 13.5 &  1.3 \\
\bottomrule
\end{tabular}%
}
\vspace{+0.2em}
\caption{\textbf{Main diagnostic results on \triad{}.}
The table reports the full diagnostic matrix used in our analysis: full
text--image--audio performance under Standard, Chain-of-thought, and
Open-ended prompts, plus group accuracy under no-input, single-modality, and
leave-one-out controls. Query accuracy measures isolated answers, while group
accuracy measures consistency across sister queries.}
\vspace{-3em}
\label{tab:main}
\end{table}

Table~\ref{tab:main} is the core evidence in the section. It reports full-input
Standard, Chain-of-thought, and Open-ended performance, together with no-input,
single-modality, and leave-one-out group accuracy. The matrix separates three
phenomena that are otherwise easy to conflate: answering isolated multiple-choice
queries, resolving a whole sister group consistently, and relying on the intended
modal evidence. We use these columns not as independent leaderboard variants,
but as probes of how a model forms and revises an interpretation when the same
scene can support multiple meanings.

\paragraph{Query accuracy overstates scene-level disambiguation.}
Under the full triplet, the strongest model reaches $90.4$ $\Accq$, close to the
$100$-query human reference at $95.1$. At the group level, however, it reaches
only $71.2$ $\Accg$ versus the human reference at $91.6$, so the reference gap
grows from $4.7$ to $20.4$ points. The
scene-group metric therefore exposes failures that isolated query accuracy hides.
The deeper issue is that many model errors are not random misses on hard items.
They indicate an unstable scene interpretation: the model can map one query to
the right option, but does not preserve the latent rule that should govern all
sisters in the group. In other words, high $\Accq$ can be achieved by local
matching, whereas high $\Accg$ requires a controllable representation of which
modality cue changes the answer.

\paragraph{Removing options exposes weak free-response resolution.}
\label{subsec:closed-vs-open}
The strongest model drops from $90.4$ to $18.4$ $\Accq$ when the option list is
hidden, and from $71.2$ to $4.6$ $\Accg$. The $100$-query human reference remains
much higher at $88.4$ $\Accq$ and $79.2$ $\Accg$. Thus multiple-choice prompting provides substantial
scaffolding: models can often select among listed interpretations without being
able to produce the resolved interpretation reliably. This suggests that the
option list is not a neutral evaluation device. It supplies a discrete hypothesis
space and lets the model rank answers by plausibility, even when the model has
not built the underlying tri-modal interpretation. The closed-to-open collapse
therefore reveals a gap between recognition among candidates and generation of
the intended meaning.

\paragraph{Ablations reveal shortcut dependence.}
Removing text hurts most, but removing image or audio leaves surprisingly high
group accuracy for strong models. For Gemini 3.1 Pro, No-Image and No-Audio
remain at $57.2$ and $51.8$ $\Accg$, even though the benchmark construction
requires all three modalities. These scores indicate that models often exploit
textual cues, partial visual evidence, and option priors instead of treating
text, image, and audio as joint constraints. The asymmetry also suggests a
specific failure mode: text often acts as the anchor that proposes an answer,
while image and audio are used as confirmatory evidence rather than as variables
that can overturn the interpretation. A strong leave-one-out score is therefore
not necessarily robustness; on \triad{}, it can mean insufficient sensitivity to
the missing modality.

\begin{figure}[t]
\centering
\begin{subfigure}[t]{0.48\linewidth}
\centering
\begin{tikzpicture}
\begin{axis}[
  width=\linewidth,
  height=3.7cm,
  ybar,
  bar width=5pt,
  ymin=0,
  ymax=100,
  ylabel={$\Accq$},
  symbolic x coords={Full,None,Text,Image,Audio,No-T,No-I,No-A},
  xtick=data,
  x tick label style={rotate=35,anchor=east,font=\scriptsize},
  yticklabel style={font=\scriptsize},
  ylabel style={font=\scriptsize},
  nodes near coords,
  nodes near coords style={font=\tiny,rotate=90,anchor=west},
  every axis plot/.append style={fill=blue!55,draw=blue!70!black},
]
\addplot coordinates {(Full,90.4) (None,29.1) (Text,85.7) (Image,46.2) (Audio,33.5) (No-T,55.4) (No-I,87.1) (No-A,82.9)};
\end{axis}
\end{tikzpicture}
\caption{Standard-prompt ablation for Gemini 3.1 Pro.}
\label{fig:ablation-pattern}
\end{subfigure}\hfill
\begin{subfigure}[t]{0.48\linewidth}
\centering
\begin{tikzpicture}
\begin{axis}[
  width=\linewidth,
  height=3.7cm,
  ymin=0,
  ymax=75,
  ylabel={Accuracy},
  xlabel={Group size},
  symbolic x coords={2,3,4,8},
  xtick=data,
  legend style={font=\scriptsize,at={(0.03,0.97)},anchor=north west,draw=none,fill=none},
  x tick label style={font=\scriptsize},
  yticklabel style={font=\scriptsize},
  xlabel style={font=\scriptsize},
  ylabel style={font=\scriptsize},
  line width=0.8pt,
  mark size=1.8pt,
]
\addplot+[blue!70!black,mark=*] coordinates {(2,70.5) (3,70.1) (4,69.6) (8,68.0)};
\addplot+[orange!85!black,mark=square*] coordinates {(2,51.8) (3,38.7) (4,27.3) (8,8.9)};
\legend{Per-query $\Accq$, Group $\Accg$}
\end{axis}
\end{tikzpicture}
\caption{Consistency collapses as sister groups grow.}
\label{fig:group-consistency}
\end{subfigure}
\vspace{-0.3em}
\caption{Two compact diagnostics kept in the main text because they support the
main claims beyond Table~\ref{tab:main}. Left: the strongest model remains high
under Text-only and leave-one-out views, revealing shortcut dependence. Right:
per-query accuracy is nearly stable across group sizes, but group accuracy drops
sharply. Full tables appear in Appendix~\ref{app:full-results}.}
\label{fig:compact-diagnostics}
\vspace{-1em}
\end{figure}
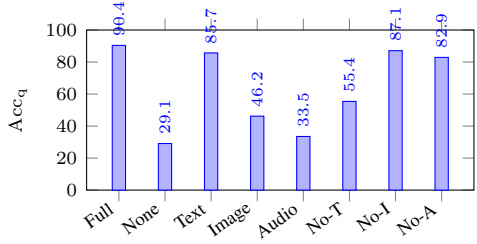
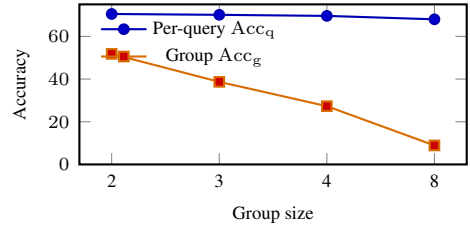

\begin{table}[t]
\centering

\setlength{\tabcolsep}{3.4pt}
\renewcommand{\arraystretch}{0.96}
\scriptsize
\resizebox{0.76\textwidth}{!}{%
\begin{tabular}{l ccc | cccccc}
\toprule
 & \multicolumn{3}{c|}{\textbf{Ambiguity tier}} & \multicolumn{6}{c}{\textbf{Selected ablation probes}} \\
\cmidrule(lr){2-4}\cmidrule(lr){5-10}
Model & 1-amb. & 2-amb. & 3-amb. & Full & None & Text & No-T & No-I & No-A \\
\midrule
Gemini 3.1 Pro       & 95.4 & 94.1 & 87.8 & 90.4 & 29.1 & 85.7 & 55.4 & 87.1 & 82.9 \\
Gemini 3 Pro         & 94.8 & 92.6 & 86.2 & 89.2 & 28.7 & 83.9 & 53.6 & 86.1 & 81.4 \\
Gemini 3 Flash       & 87.9 & 87.7 & 80.6 & 82.5 & 27.4 & 81.1 & 43.6 & 82.1 & 82.0 \\
Qwen3.5-Omni-Plus    & 93.9 & 82.3 & 74.4 & 78.0 & 21.1 & 74.5 & 43.8 & 77.5 & 75.5 \\
GPT-4o               & 87.5 & 80.9 & 72.1 & 76.3 & 25.9 & 69.7 & 44.3 & 74.0 & 70.8 \\
Gemini 2.5 Flash     & 86.9 & 81.8 & 68.6 & 73.2 & 17.4 & 69.5 & 39.6 & 74.8 & 73.0 \\
MiniCPM-o-4.5        & 78.6 & 71.4 & 61.5 & 66.2 & 22.4 & 60.1 & 36.7 & 65.0 & 61.9 \\
Qwen2.5-Omni-7B      & 73.2 & 66.8 & 57.3 & 61.8 & 20.5 & 56.4 & 33.1 & 60.2 & 58.5 \\
Baichuan-Omni-1.5    & 65.4 & 58.2 & 48.7 & 52.9 & 22.1 & 48.4 & 28.2 & 52.3 & 49.8 \\
\midrule
Human                & 97.8 & 96.2 & 93.6 & 95.1 & 24.3 & 41.8 & 46.7 & 72.5 & 68.3 \\
\midrule
Share of items       & 70.3\% & 18.0\% & 11.6\% & \multicolumn{6}{c}{--} \\
\bottomrule
\end{tabular}%
}
\vspace{+0.3em}
\caption{\textbf{Model-aligned Standard-prompt diagnostics.}
The first block reports $\Accq$ stratified by the number of individually
ambiguous modalities; a $k$-ambiguous item has $k$ modalities that are each
individually ambiguous, while the full triplet resolves the answer. The second
block reuses the same model rows to report selected query-level ablation probes:
Full input, no input, text only, and leave-one-out removal of text, image, or
audio.}
\vspace{-3em}
\label{tab:ambiguity-pattern}
\end{table}

\subsection{What explains the failures?}
\label{subsec:modality-ablation}
\label{subsec:errors}

Figure~\ref{fig:compact-diagnostics} and Table~\ref{tab:ambiguity-pattern}
highlight the auxiliary diagnostics that are most important for the main text.
First, the ablation pattern is not a standard perception result. Since every
item requires all three modalities by construction, high Text-only, No-Image, or
No-Audio scores mean that the model is using shortcuts available in the
remaining evidence and option set. The selected ablation columns in
Table~\ref{tab:ambiguity-pattern}, and the full matrix in
Table~\ref{tab:modality-ablation}, show that this pattern holds across model
families. More importantly, the direction of the shortcuts is systematic rather
than accidental. Models are much closer to Full performance when text is present
than when only image or audio is present, and removing audio often changes little
at the query level. This is the opposite of what a fully ambiguity-aware solver
should do: if audio is the cue that distinguishes two sisters, then removing it
should leave the interpretation underdetermined, not merely slightly noisier.

Second, the ambiguity-tier columns of Table~\ref{tab:ambiguity-pattern} show that performance tracks the
amount of ambiguity that must be resolved across modalities. For every model in
the table, accuracy is lowest on 3-ambiguous items; for example, Gemini 3.1 Pro
drops from $95.4$ on 1-ambiguous items to $87.8$ on 3-ambiguous items, while
humans remain above $93$ in all tiers. This makes the table a core diagnostic
rather than a secondary breakdown: the hardest cases are precisely those where
uncertainty is distributed across all three channels. The implication is that
\triad{} is not merely collecting perceptually difficult examples. It increases
difficulty when the model must maintain several partially compatible hypotheses
and use cross-modal constraints to eliminate them. Current models improve when a
single dominant cue is available, but degrade when no modality can be treated as
a self-sufficient answer source.

Third, sister consistency is the main reason we report $\Accg$. In
Figure~\ref{fig:group-consistency}, per-query accuracy changes only slightly as
group size grows from $2$ to $8$ ($70.5$ to $68.0$), but group accuracy falls
from $51.8$ to $8.9$. The model is therefore not simply worse on larger groups; it
fails to apply the same disambiguation rule consistently across controlled
sisters. This pattern points to a failure of counterfactual control. The model
can answer many individual variants, but it does not reliably understand that a
specific change in prosody, visual context, or wording should be the reason the
answer changes. Group accuracy penalizes exactly this missing invariance.

Additional breakdowns support the same interpretation. Textualizing the image or
audio recovers only part of the Full-input score, so audio is not merely a
transcript channel and visual evidence is not fully captured by captions
(Table~\ref{tab:description-ablation}). Manual error analysis further shows that
transcript-only listening, image-anchored commitment, option priors, and
cross-modal contradiction all reflect the same behavior: models often notice a
cue but commit before using all modalities to narrow the answer
(Table~\ref{tab:error-analysis}). These failures are different surface forms of
premature commitment. The model forms an early hypothesis from the easiest or
most language-like signal, then uses the remaining modalities to justify that
hypothesis instead of letting them revise it.

\subsection{Takeaway}

The experiments show that current omni-modal models are not failing only at
recognizing isolated objects, words, or sounds. They often perceive pieces of the
evidence, but do not reliably let text, image, and audio jointly determine one
interpretation. This explains the gap between $\Accq$ and $\Accg$, the collapse
from closed to open answering, the high ablated scores, and the sharp drop in
sister-group consistency. More broadly, the results suggest that present systems
still behave more like strong candidate selectors than stable tri-modal
reasoners: they can rank plausible answers when the choices are visible, but they
struggle to maintain and update a single interpretation as controlled modality
cues change.

\section{Conclusion}
\label{sec:conclusion}

\triad{} introduces a controlled diagnostic for cross-modal ambiguity, with an $18$-category taxonomy, $327$ queries in $86$ scene groups, and evaluations of $14$ omni-modal systems. Because each item is answerable only from the full text--image--audio triplet, the benchmark tests whether models jointly constrain interpretation rather than rely on a single cue. The results show a consistent gap from humans: models perform much better with closed options than open-ended answers, lose sister-group consistency, and often retain high scores under ablations, suggesting premature commitment instead of stable tri-modal reasoning.

\paragraph{Limitations.}
\triad{} is a diagnostic benchmark, not a large training corpus. Its scale leaves some strata underrepresented, especially fully three-way ambiguous items; its language and audio coverage remain narrower than real-world multimodal use; and both well-posedness checks and open-ended grading still depend on human or judge-model decisions. Future work should expand the dataset across languages, cultures, and noisier audio settings while strengthening human-centered validation.

% --- References ---
\bibliographystyle{unsrt}
\bibliography{ref}

% =====================================================================
% Appendix
% =====================================================================
\appendix
\section{Full tri-modal ambiguity taxonomy}
\label{app:taxonomy}

This appendix defines each leaf category of the \triad{} taxonomy and provides one canonical example per category. Notation follows Section~\ref{sec:benchmark-wellposedness}: $(\mathcal{C}, \mathcal{I}, \mathcal{A}, \mathcal{Q}, \mathcal{Y})$. For the worked examples below we describe the audio in [brackets] and the image in \{braces\}; the full media files are included in the released benchmark.

\subsection{Text ambiguity (6 categories)}

\paragraph{Anaphoric.} A pronoun, demonstrative, or reference target does not resolve uniquely from the sentence alone.
\emph{Example.} $\mathcal{C}$: \emph{``She told him she was leaving.''} $\mathcal{Q}$: who is leaving? $\mathcal{Y}$: \{(A) she, (B) he, (C) a third person\}.

\paragraph{Syntactic.} The sentence admits more than one grammatical parse, for example PP-attachment or coordination-scope ambiguity.
\emph{Example.} $\mathcal{C}$: \emph{``I saw the man with the telescope.''} $\mathcal{Q}$: who has the telescope? $\mathcal{Y}$: \{(A) the speaker, (B) the man, (C) someone else\}.

\paragraph{Logical.} The scope of a quantifier, a negation, or a modal is under-determined.
\emph{Example.} $\mathcal{C}$: \emph{``Every student did not finish the exam.''} $\mathcal{Q}$: which reading is intended? $\mathcal{Y}$: \{(A) none finished, (B) not all finished, (C) cannot tell\}.

\paragraph{Pragmatic.} The speech act or the implicature, not the literal content, carries the intended meaning; irony, indirect requests, and politeness reversals sit here.
\emph{Example.} $\mathcal{C}$: \emph{``Could you pass the salt?''} $\mathcal{Q}$: what is the speaker asking? $\mathcal{Y}$: \{(A) a literal ability question, (B) a polite request to pass the salt, (C) a sarcastic complaint\}.

\paragraph{Temporal.} The temporal order, tense, or aspect of the described event is under-determined by the sentence alone.
\emph{Example.} $\mathcal{C}$: \emph{``She was reaching for the door when I entered.''} $\mathcal{Q}$: was the door already open? $\mathcal{Y}$: \{(A) yes, (B) no, (C) cannot tell\}.

\paragraph{Lexical.} A content word has two or more senses (polysemy or homonymy) that the rest of the sentence does not disambiguate.
\emph{Example.} $\mathcal{C}$: \emph{``He couldn't see the bank from where he was standing.''} $\mathcal{Q}$: what kind of bank? $\mathcal{Y}$: \{(A) the bank of a river, (B) a financial institution, (C) a row of switches\}.

\subsection{Image ambiguity (6 categories)}

\paragraph{Semiotic.} A symbolic or emblematic sign admits more than one reading, for example a gesture or a cultural icon that means different things in different contexts.

\paragraph{Grounding.} Multiple candidate objects in the scene fit the question stem, so which object the question is about is itself unclear.

\paragraph{Spatial.} The relative position, orientation, or depth of objects is ambiguous from the frame alone.

\paragraph{Kinematic.} A snapshot of a dynamic event does not reveal the direction or stage of motion---door opening versus closing, plate filling versus emptying.

\paragraph{Observational.} The viewpoint or the framing makes the same physical layout compatible with different scene readings, for example a silhouette that could belong to several objects.

\paragraph{Contextual.} The scenario that the image depicts is under-determined by the pixels alone: the same whiteboard could be a classroom, a startup office, or an interview room.

\subsection{Audio ambiguity (6 categories)}

\paragraph{Referential.} The utterance contains a referring expression (pronoun, proper name, kinship term) whose referent depends on who is speaking, to whom, or in what context.

\paragraph{Prosodic.} The lexical content is held fixed across sisters while focal stress or intonation changes meaning. \emph{Example.} \emph{``I didn't say she stole the money.''} shifts interpretation depending on which of the five content words carries focal stress.

\paragraph{Indexical.} The speaker's age, gender, or social role carries information that the transcript does not. \emph{Example.} ``my boss called'' said in a child's voice versus an adult voice.

\paragraph{Phonetic.} Confusable segments, junctures, or minimal pairs (for example ``ice cream'' versus ``I scream'') resolve only with the acoustic signal or a richer context.

\paragraph{Environmental.} A background sound changes the reading of the foreground speech, for example the same sentence uttered over a hospital PA versus in a kitchen.

\paragraph{Symbolic.} A non-linguistic auditory symbol (an alarm, a doorbell, a phone ring with a recognisable ringtone) carries ambiguity-resolving information.

\subsection{Worked tri-modal example}
\label{app:taxonomy-worked}

We work one extended example to make well-posedness concrete; the released benchmark contains analogous constructions for every leaf.

\textit{Scene.} Two colleagues at a kitchen table with a pile of burnt toast between them.

\begin{itemize}[leftmargin=*,topsep=2pt,itemsep=1pt]
  \item $\mathcal{C}$: \emph{``Sure, that was a really helpful suggestion.''} (text leaf: \emph{pragmatic} --- sincere versus sarcastic reading).
  \item $\mathcal{I}$: \{two adults at the table; one smiling warmly, or rolling eyes\} (image leaf: \emph{semiotic} --- facial expression).
  \item $\mathcal{A}$: [the same sentence, recorded with sincere or sarcastic prosody] (audio leaf: \emph{prosodic}).
  \item $\mathcal{Q}$: \emph{how does the speaker feel about the suggestion?}, $\mathcal{Y}$: \{grateful, sarcastic, ambivalent, surprised\}.
\end{itemize}

Removing audio, the text and image alone leave the sincere versus sarcastic reading open when the speaker's face is neutral across sisters. Removing text, the image and audio alone cannot tell what suggestion is being evaluated or whose it was. Removing image, the text and audio alone do not anchor whether the addressee actually reacted. Only the full triplet resolves a unique answer per sister.

\subsection{Per-leaf item counts}
\label{app:taxonomy-counts}

Every \triad{} item carries one leaf label per modality, so each modality-specific leaf distribution sums to $327$. The leaf-level counts under the $327$-query release are: text---anaphoric $66$, syntactic $64$, logical $61$, pragmatic $51$, temporal $48$, lexical $37$; image---semiotic $71$, grounding $65$, spatial $61$, kinematic $53$, observational $41$, contextual $36$; audio---referential $98$, prosodic $84$, indexical $54$, phonetic $37$, environmental $34$, symbolic $20$. Figure~\ref{fig:donut} renders the same counts on the outer ring of the four-ring diagram.
\section{Audio collection and recording protocol}
\label{app:audio}

\triad{}'s audio is the most novel and the most fragile part of the benchmark. This appendix documents the protocol so that the dataset is reproducible and so that community contributors can extend it under the same standards.

\subsection{Hard rules}
\label{app:audio-rules}

\begin{enumerate}[leftmargin=*,topsep=2pt,itemsep=2pt]
  \item \textbf{No TTS-only audio.} A clip generated by a text-to-speech system without any human modification is never admitted to the benchmark. TTS audio with manual prosody editing followed by a human-naturalness check is admissible only when no human or natural recording is feasible.
  \item \textbf{No biometric or identifying content.} Voice actors sign a release covering benchmark distribution. We do not include any clip that names a specific real-world individual or that could be used to impersonate one.
  \item \textbf{Audio must be necessary.} Any clip whose contribution is purely lexical (i.e., a transcript suffices) is not \triad{} audio---by Definition~\ref{def:wellposed} the audio must contribute information beyond what the transcript carries.
  \item \textbf{No content that violates ethics rules.} Same as \S\ref{sec:construction}: no violence, privacy violations, or sensitive personal information.
\end{enumerate}

\subsection{Sources}
\label{app:audio-sources}

In decreasing order of preference:

\textbf{(S1) Natural recordings.} Permissively-licensed sources (LibriVox, Common Voice, Free Sound) used when a clip with the required prosodic feature already exists in a usable form. Provenance is recorded; for Common Voice we restrict to the \emph{validated} subset. About a third of our audio falls in this bucket.

\textbf{(S2) Voice acting.} The default source for audio that requires a specific prosodic or indexical feature. We retained four trained voice actors (two men, two women, age range $25$--$45$, all with English fluency at native or near-native level) and two Mandarin voice actors (one man, one woman, native speakers from northern and southern China respectively). For each clip the actor is given (i) the lexical content, (ii) the target prosodic feature (e.g., ``stress on the word \textsc{she}'' or ``ironic praise''), and (iii) a brief scene description for context. Actors record at least three takes and the producer, one of the authors, selects the most natural take. Cross-actor consistency is spot-checked on a $5$ percent sample by an independent listener. About three fifths of our audio comes from this bucket.

\textbf{(S3) TTS with manual prosody editing.} Used only when (S1) and (S2) fail; in our pilots this was fewer than five percent of clips. We start from a commercial neural TTS voice as the base, then manually shift formants, $F_0$ contour and duration to produce the target prosodic pattern. A naturalness check (\S\ref{app:audio-check}) is required.

\subsection{Recording specification}
\label{app:audio-spec}

All voice-acted recordings are made with the following spec: sample rate $48$\,kHz, $16$-bit PCM, mono, RMS amplitude normalised to $-23$\,LUFS. Recording is done in a treated room ($T_{60} < 0.3$\,s) with a large-diaphragm condenser microphone at $20$--$30$\,cm. Clips are kept as short as possible while preserving the ambiguity-relevant cue. We do not impose a fixed six-second cap, because some environmental, dialogue-based, and symbolic cues require longer context. Files are saved as \texttt{wav} (lossless).

\subsection{Naturalness check}
\label{app:audio-check}

Every clip is rated on a five-point Likert scale by two listeners not involved in recording, on three axes: naturalness, prosodic clarity (does the target feature come through unambiguously), and absence of artefacts. Clips with a mean score below $3.5$ on any axis are re-recorded.

\subsection{Sister-clip control within audio-perturbed groups}
\label{app:audio-sister}

For audio-perturbed groups (e.g., a stress-shift group with four sisters that move focal stress to different words) we require all sisters to be recorded by the same actor in the same session, on the same equipment, with cross-take amplitude matching. This minimises spurious cues from acoustic differences unrelated to the prosodic feature under test. We document this control in the per-group metadata.

\section{Annotator guidelines and inter-annotator agreement}
\label{app:annotator}

\subsection{Annotator pool}

Five annotators participated in \triad{} construction: three NLP graduate students (English proficiency C1+), two with formal linguistics training (English C2 or native). All annotators were familiar with the prior CODIS and MUCAR benchmarks. Each annotator was paid a market-rate stipend per query authored.

\subsection{Author--reviewer flow}

Each query goes through the following stages:
\begin{enumerate}[leftmargin=*,topsep=2pt,itemsep=2pt]
  \item \textbf{Authoring.} Annotator $A$ proposes a tuple $(\mathcal{C}, \mathcal{I}, \mathcal{A}, \mathcal{Q}, \mathcal{Y}, y^\star)$ and tags the ambiguity-modality pattern and a leaf label in each of the three taxonomies (Appendix~\ref{app:taxonomy}).
  \item \textbf{Grammar pass.} A second annotator $B$ checks the text for grammaticality and idiomaticity. Sentences are revised until both reviewers sign off.
  \item \textbf{Audio sign-off.} A third annotator $C$ listens to the audio and scores it on the naturalness axes (\S\ref{app:audio-check}). Failed clips are re-recorded.
  \item \textbf{Modality-subset checklist.} Annotators $B$, $C$, $D$ jointly walk through every strict subset $S \subsetneq \mathcal{M}$ following \S\ref{subsec:verifier}, recording, for each subset, whether at least two options remain consistent with $x_S$. A candidate is held only if every $|S|=1$ and $|S|=2$ leaves $\ge 2$ consistent options and removing any single modality enlarges the consistent option set; otherwise the candidate is returned to the author.
  \item \textbf{Independent answer check.} A fourth annotator $D$, blind to the labelled answer, predicts the answer given the full triplet. If $D$ disagrees with the labelled answer, the query is revised or discarded.
  \item \textbf{Group integration.} Annotator $E$ checks that the new query, taken together with its group sisters, satisfies the group-level constraints (single perturbed modality, consistent answer space and option semantics, distinct answers).
\end{enumerate}

A query is admitted to \triad{} only if every stage passes. Pooled across the whole pipeline, about forty percent of candidate queries are admitted; roughly a quarter fail at the modality-subset checklist stage and are either revised or discarded before the rest of the flow.

\subsection{Inter-annotator agreement}

We measured IAA on three labels, all on a held-out fifty-query audit subset:

\begin{itemize}[leftmargin=*,topsep=2pt,itemsep=2pt]
  \item Correctness of $y^\star$: Cohen's $\kappa = 0.84$ across five annotator pairs.
  \item Per-modality ambiguity tier tag: $\kappa = 0.79$. Disagreements concentrate on the boundary between \emph{temporal} and \emph{logical} text items, and on the boundary between \emph{prosodic} and \emph{indexical} audio items.
  \item Naturalness of audio (binary admit / re-record): $\kappa = 0.81$.
\end{itemize}

Per-leaf IAA and the disagreement matrix are released together with the dataset.

\subsection{Reviewer guidance for degenerate ambiguities}

Annotators were specifically trained on the failure modes that we want \emph{not} to admit:

\begin{itemize}[leftmargin=*,topsep=2pt,itemsep=2pt]
  \item \textbf{Reducible items.} If $(\mathcal{C}, \mathcal{I}, \mathcal{Q}, \mathcal{Y})$ already determines $y^\star$, the audio is decorative; reject. Symmetric rule for the other two-modality slices.
  \item \textbf{Trick options.} If one option in $\mathcal{Y}$ is manifestly false to a literate reader regardless of all modalities, replace it with a plausible distractor.
  \item \textbf{Semantic leak through option strings.} If the option strings embed modality-specific evidence (e.g., ``the man on the left'' when there is only a left and a right person), this leaks the answer to text-only models; reword the options to be modality-neutral wherever possible.
\end{itemize}

\section{Well-posedness: formal treatment and reference checker}
\label{app:irreducibility}

\subsection{Formal restatement}

We restate Definition~\ref{def:wellposed} with the non-degeneracy side condition made explicit. For a query $x = (\mathcal{C}, \mathcal{I}, \mathcal{A}, \mathcal{Q}, \mathcal{Y})$ and a modality subset $S \subseteq \mathcal{M} = \{\mathcal{T}, \mathcal{V}, \mathcal{A}\}$, let
\[
  \mathcal{Y}_S(x) = \{y \in \mathcal{Y} : y \text{ is consistent with } x_S\},
\]
where $x_S$ is the input restricted to modalities in $S$ (the question $\mathcal{Q}$ and option set $\mathcal{Y}$ are always provided). $x$ is \emph{well-posed} iff
\begin{align}
  |\mathcal{Y}_{\mathcal{M}}(x)| &= 1, \\
  |\mathcal{Y}_{\mathcal{M} \setminus \{m\}}(x)| &> 1 \quad \text{for every } m \in \mathcal{M}.
\end{align}
The human checklist also asks reviewers to inspect every strict subset of modalities. Under the usual monotonic reading of consistency, if each leave one out view has more than one consistent option, then single modality views cannot have fewer options. We still include the single modality check because it makes the review procedure concrete and catches annotation mistakes where a question or option string leaks the answer.

\subsection{Reference well-posedness checker}
\label{app:irreducibility-verifier}

We release a stand-alone Python implementation of a modality-subset probe that approximates $\mathcal{Y}_S(x)$ by querying omni-modal LLMs under a forced-choice prompt (Appendix~\ref{app:prompts}) and aggregating across $K$ samples at temperature $0.7$:
\[
  \widehat{\mathcal{Y}}_S(x) = \bigl\{ y \in \mathcal{Y} :
     \text{either configured probe selects $y$ in at least $\tau \cdot K$ of the $K$ samples} \bigr\}.
\]
The default configuration ships with GPT-4o and Gemini~2.5~Pro as example clients, $K = 5$ for two modality views, $K = 10$ for the full triplet, and threshold $\tau = 0.20$. Under this configuration the checker flags an item as suspect when $|\widehat{\mathcal{Y}}_S(x)| < 2$ for any $|S| = 2$ or when removing some $m$ leaves $\widehat{\mathcal{Y}}$ unchanged from $\widehat{\mathcal{Y}}_{\mathcal{M}}$. The released $327$ items were selected by five annotators using the modality subset checklist (\S\ref{subsec:verifier}); the checker is shipped as a reproducibility tool and as a screening aid for downstream users, not as the gate that produced the benchmark.

The defaults are intentionally permissive: $\tau = 0.20$ requires that an option be chosen in only one of five samples to count as ``consistent'', which makes the checker sensitive to weak alternatives. We report the configuration so that users running the checker on community contributed extensions or on other omni-modal benchmarks can reproduce the same screening regime; tighter thresholds ($\tau = 0.40$ or higher) yield a stricter notion of consistency at the cost of rejecting more borderline items.

\subsection{Intended uses of the checker}

The reference checker is intended for three downstream uses. First, screening community-contributed items before they are added to a future \triad{} release. Second, applying the same well-posedness property to other omni-modal benchmarks, when the contributors of those benchmarks wish to characterise how much of their data is solvable from a strict subset of modalities. Third, providing a reproducible second opinion alongside the human admission record \triad{} ships with: each item's modality-subset probe outputs can be regenerated by running the checker against the released triplets, so users do not have to take our cross-review notes at face value. The checker is not a definition of correctness---it is a probe-based approximation to the option-consistency sets in Definition~\ref{def:wellposed}, and the labelled answer is always determined by human annotation.

\section{Supplementary experimental results}
\label{app:full-results}

This appendix collects numerical breakdowns that support the main experimental
section (Section~\ref{sec:exp}) but are not necessary for reading the central
diagnostic results in Table~\ref{tab:main} and Figure~\ref{fig:compact-diagnostics}.
The main paper keeps the full diagnostic matrix and a compact visual summary;
the tables below document detailed ablations, counterfactuals, robustness
checks, secondary stratifications, and grading reliability.

\subsection{Result provenance and aggregation}
\label{app:result-provenance}

All reported model scores are computed from the same $327$ query identifiers and
$86$ scene-group identifiers unless a table explicitly states otherwise. The
Standard and Chain-of-thought columns in Table~\ref{tab:main} come from one
answer per model--prompt--condition--query tuple at temperature $0$. For closed
multiple-choice prompts, parsing first looks for an explicit option letter; when
no unambiguous letter is present, the prediction is counted as incorrect rather
than manually repaired. The Open-ended columns use the same media inputs and the
same questions, but remove the option block before generation. Query accuracy
is the mean correctness over the relevant query set. Group accuracy is computed
by first scoring each query, then marking a group correct only if every sister
query in that group is correct; for Open-ended mean-judge scores, the group
score is the product of the per-sister mean judge scores.

The eight Standard-prompt input conditions are: Full, None, Text-only,
Image-only, Audio-only, No-Text, No-Image, and No-Audio. Table~\ref{tab:main}
reports the group-accuracy subset of these conditions, while
Table~\ref{tab:modality-ablation} reports the corresponding query-accuracy
matrix. Table~\ref{tab:description-ablation} is a separate counterfactual run on
the same $327$ queries for the six representative model endpoints and the human
row shown in the table.

\subsection{Detailed modality ablations}
\label{app:modality-ablation}

\begin{table}[t]
\centering
\small
\caption{\textbf{Modality ablation under the Standard prompt.}
Each cell reports $\Accq$ on the full $327$-query benchmark. Full uses all
three modalities; None uses only the question stem and option list;
single-modality conditions use one modality plus the question and options; and
leave-one-out conditions remove the named modality.}
\label{tab:modality-ablation}
\setlength{\tabcolsep}{3.4pt}
\renewcommand{\arraystretch}{0.98}
\begin{tabular}{l cc ccc ccc}
\toprule
 & \multicolumn{2}{c}{Global} & \multicolumn{3}{c}{Single modality} & \multicolumn{3}{c}{Leave-one-out} \\
\cmidrule(lr){2-3}\cmidrule(lr){4-6}\cmidrule(lr){7-9}
Model & Full & None & Text & Image & Audio & No-T & No-I & No-A \\
\midrule
Gemini 3.1 Pro      & 90.4 & 29.1 & 85.7 & 46.2 & 33.5 & 55.4 & 87.1 & 82.9 \\
Gemini 3 Pro        & 89.2 & 28.7 & 83.9 & 44.6 & 32.6 & 53.6 & 86.1 & 81.4 \\
Gemini 3 Flash      & 82.5 & 27.4 & 81.1 & 36.9 & 30.9 & 43.6 & 82.1 & 82.0 \\
Gemini 2.5 Pro      & 80.7 & 26.1 & 74.8 & 37.4 & 31.2 & 46.0 & 78.2 & 72.1 \\
Qwen3.5-Omni-Plus   & 78.0 & 21.1 & 74.5 & 38.3 & 28.4 & 43.8 & 77.5 & 75.5 \\
GPT-4o              & 76.3 & 25.9 & 69.7 & 40.2 & 30.1 & 44.3 & 74.0 & 70.8 \\
Gemini 2.5 Flash    & 73.2 & 17.4 & 69.5 & 30.8 & 26.9 & 39.6 & 74.8 & 73.0 \\
Qwen3-Omni-30B      & 72.5 & 20.8 & 67.3 & 34.1 & 27.6 & 40.9 & 71.8 & 68.4 \\
MiniCPM-o-4.5       & 66.2 & 22.4 & 60.1 & 32.5 & 26.3 & 36.7 & 65.0 & 61.9 \\
Qwen2.5-Omni-7B     & 61.8 & 20.5 & 56.4 & 29.8 & 25.4 & 33.1 & 60.2 & 58.5 \\
OmniVinci-9B        & 58.7 & 21.9 & 53.2 & 28.6 & 24.8 & 31.4 & 57.0 & 55.8 \\
Ola-7B              & 55.4 & 19.7 & 51.6 & 27.3 & 23.9 & 29.8 & 54.1 & 52.7 \\
Qwen2.5-Omni-3B     & 54.1 & 19.0 & 49.8 & 26.5 & 24.1 & 28.6 & 53.6 & 51.0 \\
Baichuan-Omni-1.5   & 52.9 & 22.1 & 48.4 & 26.0 & 23.7 & 28.2 & 52.3 & 49.8 \\
\midrule
Human               & 95.1 & 24.3 & 41.8 & 38.6 & 35.2 & 46.7 & 72.5 & 68.3 \\
\bottomrule
\end{tabular}
\end{table}

Table~\ref{tab:modality-ablation} gives the full Standard-prompt query-accuracy
matrix for the input conditions summarized in the main text. It separates
option-prior behavior under the None condition, single-modality behavior, and
leave-one-out behavior.

\subsection{Description-ablation protocol}
\label{app:description-ablation-protocol}

For the audio-description condition A$^+$, annotators wrote a compact note that
contains the transcript, prosody relevant to disambiguation, speaker or voice
properties when needed, and non-speech acoustic events. The note is not allowed
to mention the answer option, the leaf category, or any sister identifier. For
the image-description condition I$^+$, annotators wrote a literal scene
caption covering objects, agents, spatial relations, visible text, gestures, and
motion cues when inferable from the image. Descriptions are written once per
item and then used unchanged for all evaluated models. This design tests whether
a human textualization of a modality preserves the disambiguating signal; it is
not intended as an optimized prompt-engineering condition.

\begin{table}[t]
\centering
\small
\caption{Replacing non-text modalities with human-written descriptions. Each cell reports $\Accq$ under the Standard prompt on the same $327$ queries. A$^+$ denotes audio replaced by a transcript with prosody and non-speech-event notes; I$^+$ denotes image replaced by a human-written scene description. Numbers in parentheses are deltas from the model's Full score in Table~\ref{tab:modality-ablation}.}
\label{tab:description-ablation}
\setlength{\tabcolsep}{5pt}
\begin{tabular}{l c c c c}
\toprule
Model & Full (T+I+A) & T+I+A+ & T+I$^+$+A & T+I$^+$+A+ \\
\midrule
Gemini 3 Pro      & 89.2 & 83.6 ($-5.6$) & 85.4 ($-3.8$) & 79.8 ($-9.4$) \\
Gemini 3 Flash    & 82.5 & 77.1 ($-5.4$) & 79.3 ($-3.2$) & 74.0 ($-8.5$) \\
Qwen3.5-Omni-Plus & 78.0 & 71.4 ($-6.6$) & 74.2 ($-3.8$) & 68.1 ($-9.9$) \\
GPT-4o            & 76.3 & 69.9 ($-6.4$) & 72.8 ($-3.5$) & 65.7 ($-10.6$) \\
Gemini 2.5 Flash  & 73.2 & 66.8 ($-6.4$) & 69.5 ($-3.7$) & 62.3 ($-10.9$) \\
MiniCPM-o-4.5     & 66.2 & 58.9 ($-7.3$) & 62.0 ($-4.2$) & 54.7 ($-11.5$) \\
\midrule
Human             & 95.1 & 88.7 ($-6.4$) & 90.5 ($-4.6$) & 84.1 ($-11.0$) \\
\bottomrule
\end{tabular}
\end{table}

Table~\ref{tab:description-ablation} reports the corresponding counterfactual
scores. It is used in the main text only to support the conclusion that audio and
image carry disambiguating information beyond a textual note.

\subsection{Ambiguity tiers and sister consistency}
\label{app:ambiguity-consistency}

Table~\ref{tab:ambiguity-pattern} in the main paper stratifies Standard-prompt
accuracy by the number of individually ambiguous modalities and reuses the same
model rows for selected ablation probes. Table~\ref{tab:group-size} below gives
the detailed group-size breakdown behind the sister-consistency curve in
Figure~\ref{fig:group-consistency}.

\begin{table}[t]
\centering
\small
\caption{Group-size distribution and the relationship between group size, in-group per-query accuracy, and group-wise accuracy. Per-query and $\Accg$ numbers are averaged across all 14 models under the Standard prompt.}
\label{tab:group-size}
\setlength{\tabcolsep}{8pt}
\begin{tabular}{l c c c c}
\toprule
Group size & 2 & 3 & 4 & 8 \\
\midrule
Number of groups           & 23    & 11    & 42    & 10    \\
Number of queries          & 46    & 33    & 168   & 80    \\
Share of queries           & 14.1\% & 10.1\% & 51.4\% & 24.5\% \\
Per-query $\Accq$ (in-group) & 70.5 & 70.1 & 69.6 & 68.0 \\
$\Accg$                     & 51.8 & 38.7 & 27.3 & 8.9  \\
\bottomrule
\end{tabular}
\end{table}

\subsection{Benchmark comparison details}
\label{app:bench-comparison}

% Comparison of TRIAD with prior ambiguity / omni-modal benchmarks.
\begin{table}[t]
\centering
\small
\caption{Comparison of TRIAD with prior multimodal benchmarks. ``Ambig.'' columns
mark whether the benchmark explicitly induces ambiguity in that modality.
``Irred.'' = whether the benchmark guarantees that no two-modality subquery is
sufficient. ``Group'' = whether items are organized into perturbation groups.}
\label{tab:bench-compare}
\begin{tabular}{l c c c c c c c}
\toprule
Benchmark
  & \makecell{Text\\Ambig.}
  & \makecell{Image\\Ambig.}
  & \makecell{Audio\\Ambig.}
  & \makecell{Tri-modal}
  & \makecell{Irred.}
  & \makecell{Group}
  & Evaluator \\
\midrule
MME              & \xmark & \xmark & \xmark & \xmark & \xmark & \xmark & Metrics \\
HallusionBench & \xmark & \cmark & \xmark & \xmark & \xmark & \xmark & Metrics \\
CODIS           & \xmark & \cmark & \xmark & \xmark & \xmark & \cmark & Human / GPT \\

VAGUE            & \cmark & \xmark & \xmark & \xmark & \xmark & \xmark & Metrics \\
3AM                & \cmark & \xmark & \xmark & \xmark & \xmark & \xmark & Metrics \\
MUCAR            & \cmark & \cmark & \xmark & \xmark & \cmark$^{\dagger}$ & \cmark & Human / GPT \\
\midrule
AVQA               & \xmark & \xmark & \xmark & \cmark & \xmark & \xmark & Metrics \\
MUSIC-AVQA   & \xmark & \xmark & \xmark & \cmark & \xmark & \xmark & Metrics \\
AVHBench      & \xmark & \xmark & \xmark & \cmark & \xmark & \xmark & Metrics \\
OmniBench     & \xmark & \xmark & \xmark & \cmark & \xmark & \xmark & Metrics \\
\midrule
TRIAD (Ours)                    & \cmark & \cmark & \cmark & \cmark & \cmark & \cmark & Human / GPT \\
\bottomrule
\end{tabular}
\\[2pt]
{\footnotesize $^{\dagger}$MUCAR's \emph{dual-ambiguity} slice satisfies
two-modality irreducibility; TRIAD extends this to three modalities.}
\end{table}

Table~\ref{tab:bench-compare} expands the related-work discussion by separating
whether prior benchmarks explicitly induce ambiguity in text, image, or audio,
whether they are truly tri-modal, and whether they enforce a no-two-modality
sufficiency criterion.

\subsection{Confidence intervals and significance testing}

The released result files include $95\%$ bootstrap confidence intervals
computed over \emph{groups}, not queries; sisters within a group are not
independent and a query-level bootstrap would underestimate variance. We draw
$B = 10{,}000$ resamples. For each pairwise model comparison reported in
Section~\ref{sec:exp}, we additionally run a paired bootstrap at the group level
with Holm--Bonferroni correction. Ranking differences of three points or more
on $\Accg$ are significant at the $p < 0.05$ level after correction; smaller
differences are treated as not significant.

\subsection{Human and random reference points}

\begin{table}[t]
\centering
\small
\caption{Human and random baselines versus the best model on TRIAD. Random $\Accq$ uses the expected value under uniform guessing given per-item option counts; Random $\Accg$ uses the expected product-of-reciprocal-option-counts within each group. Human scores are the mean of three raters on a cost-controlled 100-query stratified subset; the same subset is used for Standard multiple choice and Open-ended answering. Model and random rows are computed on the full 327-query benchmark unless otherwise stated, so human rows should be read as a reference rather than a paired full-benchmark comparison.}
\label{tab:human-random}
\setlength{\tabcolsep}{6pt}
\begin{tabular}{l c c c c c c}
\toprule
 & \multicolumn{2}{c}{Standard} & \multicolumn{2}{c}{CoT} & \multicolumn{2}{c}{Open-ended} \\
\cmidrule(lr){2-3} \cmidrule(lr){4-5} \cmidrule(lr){6-7}
 & $\Accq$ & $\Accg$ & $\Accq$ & $\Accg$ & $\Accq$ & $\Accg$ \\
\midrule
Random                      & 24.3 & 2.2  & --   & --   & --   & --   \\
Best model (Gemini 3.1 Pro) & 90.4 & 71.2 & 91.1 & 72.5 & 18.4 &  4.6 \\
Human                       & 95.1 & 91.6 & --   & --   & 88.4 & 79.2 \\
\midrule
Gap to human reference      & $-4.7$ & $-20.4$ & $-4.0$ & $-19.1$ & $-70.0$ & $-74.6$ \\
\bottomrule
\end{tabular}
\end{table}

Table~\ref{tab:human-random} isolates the upper and lower reference rows from
Table~\ref{tab:main}. It makes explicit that the random baseline is computed
from the empirical option counts and scene-group sizes rather than from a fixed
four-way-choice assumption.
For query accuracy, the random score is
$\frac{1}{327}\sum_i |\mathcal{Y}_i|^{-1}$. For group accuracy, the random
score is $\frac{1}{86}\sum_g \prod_{i\in g}|\mathcal{Y}_i|^{-1}$, because all
sisters in a group must be answered correctly. Human Standard and Human Open-ended
scores are measured on the same $100$-item stratified subset balanced by language,
ambiguity tier, and modality leaf; the subset is used for human-cost control and
is reported separately in the release metadata.

\subsection{Prompt robustness}
\label{app:prompt-robustness}

\begin{table}[t]
\centering
\small
\caption{Prompt robustness. 3-prompt agreement is the fraction of queries on which Standard, CoT, and Open-ended all return the same option (Open-ended is mapped to an option by lexical match to the option text; items with no match are counted as disagreement). This number is bounded above by Open-ended accuracy and is low across the pool, reflecting the multiple choice to Open-ended collapse. Std-vs-CoT agreement is the fraction of queries on which the two multiple choice prompts agree, which is a cleaner prompt sensitivity measure. Spread is the max-min range of $\Accq$ across three paraphrased Standard prompts on the Full condition.}
\label{tab:prompt-robustness}
\setlength{\tabcolsep}{2pt}
\begin{tabular}{l c c c c c c}
\toprule
Model & 3-prompt agr. [0,1] & Std-vs-CoT agr. [0,1] & Std $\Accq$ & CoT $\Accq$ & Open $\Accq$ & Std spread \\
\midrule
Gemini 3.1 Pro       & 0.17 & 0.88 & 90.4 & 91.1 & 18.4 & 1.5 \\
Gemini 3 Pro         & 0.15 & 0.87 & 89.2 & 90.0 & 16.1 & 1.9 \\
Gemini 3 Flash       & 0.16 & 0.84 & 82.5 & 82.0 & 17.6 & 1.6 \\
Qwen3.5-Omni-Plus    & 0.16 & 0.81 & 78.0 & 78.2 & 17.9 & 1.8 \\
GPT-4o               & 0.14 & 0.83 & 76.3 & 78.1 & 15.4 & 2.3 \\
Gemini 2.5 Flash     & 0.14 & 0.79 & 73.2 & 74.1 & 15.8 & 5.4 \\
Qwen3-Omni-30B       & 0.13 & 0.78 & 72.5 & 72.8 & 14.7 & 2.8 \\
MiniCPM-o-4.5        & 0.11 & 0.76 & 66.2 & 68.0 & 12.3 & 3.1 \\
Qwen2.5-Omni-7B      & 0.09 & 0.74 & 61.8 & 63.9 & 11.0 & 3.4 \\
Qwen2.5-Omni-3B      & 0.07 & 0.71 & 54.1 & 53.5 &  8.4 & 4.2 \\
Baichuan-Omni-1.5    & 0.06 & 0.69 & 52.9 & 52.4 &  7.6 & 4.6 \\
\bottomrule
\end{tabular}
\end{table}

Table~\ref{tab:prompt-robustness} shows that Standard and Chain-of-thought
answers are relatively stable compared with the much larger closed-to-open
collapse. This supports the choice to focus the main text on the Standard,
Chain-of-thought, and Open-ended columns of Table~\ref{tab:main} rather than on
small prompt-paraphrase effects.

\subsection{Open-ended grading reliability}
\label{app:judge-agreement}

\begin{table}[t]
\centering
\small
\caption{Standard multiple choice versus Open-ended accuracy. Model rows use the same $327$ queries in both settings; the human row is measured on the cost-controlled 100-query stratified subset described in Appendix~\ref{app:full-results}, using the same subset for Standard and Open-ended. Open-ended $\Accq$ is the mean of two independent judges, GPT-5.1 and Claude-4.5, grading short free-form answers where no option list is shown; Strict is the fraction of items both judges mark correct. $\Delta$ is Standard minus Open-ended. The collapse from Standard to Open-ended is an order-of-magnitude effect across the model pool, while the human reference remains substantially more robust in both query and group accuracy.}
\label{tab:closed-vs-open}
\setlength{\tabcolsep}{2pt}
\begin{tabular}{l c c c c c c c}
\toprule
 & \multicolumn{4}{c}{$\Accq$} & \multicolumn{3}{c}{$\Accg$} \\
\cmidrule(lr){2-5}\cmidrule(lr){6-8}
Model & Standard & Open (mean) & Open (strict) & $\Delta$ & Standard & Open (mean) & $\Delta$ \\
\midrule
Human                & 95.1 & 88.4 & 82.7 & $+6.7$  & 91.6 & 79.2 & $+12.4$ \\
\midrule
Gemini 3.1 Pro       & 90.4 & 18.4 &  9.4 & $+72.0$ & 71.2 & 4.6  & $+66.6$ \\
Gemini 3 Pro         & 89.2 & 16.1 &  8.8 & $+73.1$ & 68.9 & 3.1  & $+65.8$ \\
Gemini 3 Flash       & 82.5 & 17.6 &  9.2 & $+64.9$ & 59.4 & 3.5  & $+55.9$ \\
Gemini 2.5 Pro       & 80.7 & 16.9 &  8.6 & $+63.8$ & 55.3 & 3.3  & $+52.0$ \\
Qwen3.5-Omni-Plus    & 78.0 & 17.9 &  9.2 & $+60.1$ & 52.8 & 3.1  & $+49.7$ \\
GPT-4o               & 76.3 & 15.4 &  7.5 & $+60.9$ & 49.8 & 2.8  & $+47.0$ \\
Gemini 2.5 Flash     & 73.2 & 15.8 &  7.8 & $+57.4$ & 45.6 & 2.9  & $+42.7$ \\
Qwen3-Omni-30B       & 72.5 & 14.7 &  7.0 & $+57.8$ & 41.7 & 2.3  & $+39.4$ \\
MiniCPM-o-4.5        & 66.2 & 12.3 &  5.7 & $+53.9$ & 36.4 & 1.6  & $+34.8$ \\
Qwen2.5-Omni-7B      & 61.8 & 11.0 &  5.1 & $+50.8$ & 30.2 & 1.2  & $+29.0$ \\
OmniVinci-9B         & 58.7 &  9.8 &  4.3 & $+48.9$ & 26.9 & 1.0  & $+25.9$ \\
Ola-7B               & 55.4 &  9.1 &  3.9 & $+46.3$ & 23.8 & 0.9  & $+22.9$ \\
Qwen2.5-Omni-3B      & 54.1 &  8.4 &  3.6 & $+45.7$ & 22.1 & 0.7  & $+21.4$ \\
Baichuan-Omni-1.5    & 52.9 &  7.6 &  3.2 & $+45.3$ & 20.4 & 0.6  & $+19.8$ \\
\bottomrule
\end{tabular}
\end{table}

\begin{table}[t]
\centering
\small
\caption{Per-model agreement between the two Open-ended judges, GPT-5.1 and Claude-4.5, on parseable short free-form answers (no option list shown). The $N$ column counts parseable responses only; these judge-specific accuracies are used for agreement analysis and are not directly averaged to obtain the full-denominator Open-ended scores in Table~\ref{tab:main}. $\kappa$ is Cohen's kappa on the binary correct/wrong verdict. Judge agreement is the fraction of items for which the two judges return the same verdict. Human-on-disagreement is the fraction of disagreement items where the averaged judge score matches an adjudicating human grader.}
\label{tab:judge-agreement}
\setlength{\tabcolsep}{2pt}
\begin{tabular}{l c c c c c c}
\toprule
Model & $N$ & \makecell{Parseable-only\\Acc (GPT-5.1)} & \makecell{Parseable-only\\Acc (Claude-4.5)} & Agreement & $\kappa$ & \makecell{Human-on-\\disagreement} \\
\midrule
Gemini 3.1 Pro       & 308 & 24.1 & 12.7 & 0.811 & 0.573 & 0.71 \\
Gemini 3 Pro         & 306 & 20.9 & 11.3 & 0.816 & 0.562 & 0.68 \\
Gemini 3 Flash       & 316 & 22.9 & 12.3 & 0.810 & 0.574 & 0.70 \\
Qwen3.5-Omni-Plus    & 315 & 24.1 & 12.2 & 0.808 & 0.562 & 0.69 \\
Gemini 2.5 Flash     & 319 & 21.2 & 10.3 & 0.821 & 0.552 & 0.67 \\
GPT-4o               & 311 & 20.0 &  9.8 & 0.827 & 0.548 & 0.66 \\
Gemini 2.5 Pro       & 314 & 22.4 & 11.1 & 0.818 & 0.558 & 0.68 \\
Qwen3-Omni-30B       & 312 & 19.3 &  9.7 & 0.823 & 0.541 & 0.65 \\
MiniCPM-o-4.5        & 309 & 16.2 &  8.1 & 0.830 & 0.519 & 0.63 \\
Qwen2.5-Omni-7B      & 313 & 14.5 &  6.9 & 0.838 & 0.498 & 0.61 \\
OmniVinci-9B         & 310 & 12.7 &  6.2 & 0.842 & 0.481 & 0.60 \\
Ola-7B               & 308 & 11.9 &  5.7 & 0.848 & 0.468 & 0.58 \\
Qwen2.5-Omni-3B      & 311 & 10.8 &  5.1 & 0.852 & 0.449 & 0.57 \\
Baichuan-Omni-1.5    & 307 &  9.9 &  4.6 & 0.857 & 0.431 & 0.56 \\
\bottomrule
\end{tabular}
\end{table}

Table~\ref{tab:closed-vs-open} reports the strict both-judge Open-ended variant
in addition to the mean judge score used in the main table. Table~\ref{tab:judge-agreement}
then reports judge agreement and human adjudication on disagreements. These
numbers justify treating Open-ended scores as a lower-bound diagnostic rather
than as a conventional free-response leaderboard.
The $N$ column in Table~\ref{tab:judge-agreement} counts Open-ended responses
that are parseable as short answers and accepted by both judge APIs. Empty
outputs, safety refusals, and malformed multi-answer lists are scored as
incorrect in Table~\ref{tab:main}; they are excluded only from the agreement
rate denominator because agreement is undefined when a judge receives no
candidate answer to classify. These parseable-only judge statistics are used for
agreement analysis and are not directly averaged to obtain the full-denominator
Open-ended scores in Table~\ref{tab:main}.

\subsection{Secondary stratifications}

\begin{table}[t]
\centering
\small
\caption{$\Accq$ under the Standard prompt, binned by audio duration. Duration bins are broad diagnostics over the $141$ clips with available duration metadata; the released benchmark contains $143$ unique audio clips in total. They are included only to check whether long clips explain the main failure pattern.}
\label{tab:duration-analysis}
\setlength{\tabcolsep}{8pt}
\begin{tabular}{l c c c}
\toprule
Model & Short ($<5$s) & Medium (5--10s) & Long ($>10$s) \\
\midrule
Gemini 3 Pro         & 90.1 & 88.7 & 88.4 \\
Gemini 3 Flash       & 83.2 & 82.0 & 81.9 \\
Qwen3.5-Omni-Plus    & 79.3 & 77.8 & 76.4 \\
GPT-4o               & 77.1 & 76.0 & 75.2 \\
Gemini 2.5 Flash     & 74.5 & 72.8 & 71.6 \\
MiniCPM-o-4.5        & 67.2 & 65.9 & 65.1 \\
Qwen2.5-Omni-7B      & 62.8 & 61.3 & 60.9 \\
Baichuan-Omni-1.5    & 53.6 & 52.5 & 52.1 \\
\midrule
Human                & 95.4 & 95.2 & 94.6 \\
\midrule
Share of items       & \multicolumn{3}{c}{reported in release metadata} \\
\bottomrule
\end{tabular}
\end{table}

\begin{table}[t]
\centering
\small
\caption{Representative leaf-level $\Accq$ under the Standard prompt. Columns are
selected leaves from the TRIAD taxonomy; the full $14 \times 18$ leaf table is
released in machine-readable form. Lower audio prosody and symbolic scores
match the main finding that acoustic cues are harder to use as ambiguity
resolvers than text or image cues.}
\label{tab:per-category}
\setlength{\tabcolsep}{3pt}
\begin{tabular}{l ccc ccc ccc}
\toprule
 & \multicolumn{3}{c}{Text leaves} & \multicolumn{3}{c}{Image leaves} & \multicolumn{3}{c}{Audio leaves} \\
\cmidrule(lr){2-4}\cmidrule(lr){5-7}\cmidrule(lr){8-10}
Model & Prag. & Temp. & Synt. & Semiot. & Kinem. & Ground. & Prosod. & Symbol. & Refer. \\
\midrule
Gemini 3.1 Pro    & 87.3 & 89.0 & 93.1 & 84.2 & 86.7 & 92.8 & 79.6 & 81.7 & 88.1 \\
Gemini 3 Pro      & 85.6 & 87.4 & 91.7 & 82.8 & 84.9 & 91.4 & 77.4 & 79.8 & 86.9 \\
Gemini 3 Flash    & 77.4 & 79.5 & 85.0 & 75.1 & 77.8 & 85.6 & 68.2 & 72.1 & 79.5 \\
Qwen3.5-Omni-Plus & 73.2 & 75.1 & 82.7 & 71.8 & 74.0 & 82.4 & 65.8 & 70.4 & 77.9 \\
GPT-4o            & 71.6 & 73.2 & 80.5 & 70.4 & 72.5 & 81.2 & 61.3 & 66.7 & 73.8 \\
Gemini 2.5 Flash  & 67.1 & 69.0 & 76.8 & 66.8 & 68.4 & 77.5 & 58.4 & 62.5 & 69.2 \\
MiniCPM-o-4.5     & 60.5 & 62.8 & 71.4 & 61.3 & 63.0 & 71.8 & 52.1 & 56.4 & 63.7 \\
Qwen2.5-Omni-7B   & 56.2 & 58.1 & 65.2 & 55.7 & 57.0 & 66.5 & 48.3 & 51.2 & 58.1 \\
Qwen2.5-Omni-3B   & 48.1 & 50.3 & 57.3 & 48.6 & 50.1 & 58.9 & 41.7 & 44.9 & 51.2 \\
Baichuan-Omni-1.5 & 46.7 & 48.8 & 56.2 & 47.4 & 49.0 & 57.1 & 40.1 & 43.3 & 49.5 \\
\midrule
Human             & 94.3 & 95.1 & 96.4 & 93.1 & 94.0 & 97.2 & 92.4 & 92.9 & 94.7 \\
\bottomrule
\end{tabular}
\end{table}

\begin{table}[t]
\centering
\small
\caption{Scaling within two model families on TRIAD. $\Accg$ under the Standard prompt. Parameter counts for the proprietary models are not disclosed; we list them by release tier.}
\label{tab:scaling}
\setlength{\tabcolsep}{8pt}
\begin{tabular}{l l c c}
\toprule
Family & Model & Size / Tier & $\Accg$ \\
\midrule
\multirow{4}{*}{Qwen omni}
  & Qwen2.5-Omni-3B      & 3B        & 22.1 \\
  & Qwen2.5-Omni-7B      & 7B        & 30.2 \\
  & Qwen3-Omni-30B       & 30B       & 41.7 \\
  & Qwen3.5-Omni-Plus    & Plus tier & 52.8 \\
\midrule
\multirow{4}{*}{Gemini}
  & Gemini 2.5 Flash     & Flash tier & 45.6 \\
  & Gemini 3 Flash       & Flash tier & 59.4 \\
  & Gemini 3 Pro         & Pro tier   & 68.9 \\
  & Gemini 3.1 Pro       & Pro tier   & 71.2 \\
\midrule
Human & --               & --         & 91.6 \\
\bottomrule
\end{tabular}
\end{table}

Duration, category, and scale analyses are useful for interpreting the main
failure pattern but are less central than the full-input, ablation, and group
consistency results. Table~\ref{tab:duration-analysis} shows that audio duration
does not explain the difficulty. Table~\ref{tab:per-category} summarizes the
hardest representative categories, with the full $14 \times 18$ leaf table
released in machine-readable form. Table~\ref{tab:scaling} shows that scaling
improves $\Accg$ within model families but does not close the human gap.

\subsection{Manual error-analysis protocol}
\label{app:error-protocol}

\begin{table}[t]
\centering
\small
\caption{Manual categorization of 200 sampled incorrect predictions, pooled across models and prompts. Shares are out of the 200 sampled errors.}
\label{tab:error-analysis}
\setlength{\tabcolsep}{8pt}
\begin{tabular}{l c l}
\toprule
Category & Share & Typical trigger \\
\midrule
Transcript-only listening   & 35.0\% & Audio treated as text; prosody ignored \\
Image-anchored commitment   & 25.5\% & Early decision from image, post-hoc justification \\
Option priors               & 20.0\% & Answer from question stem alone \\
Cross-modal contradiction   & 19.5\% & Text and audio disagree; model sides with text \\
\bottomrule
\end{tabular}
\end{table}

The $200$ errors in Table~\ref{tab:error-analysis} are sampled from incorrect
model predictions after stratifying by model family, prompt format, and
ambiguity tier. Two annotators independently assign one primary failure label
using the definitions in the table caption and the surrounding text in
Section~\ref{subsec:errors}. Disagreements are resolved by discussion; if an
error plausibly has multiple causes, the label records the earliest commitment
that makes the final answer wrong. The percentages in Table~\ref{tab:error-analysis}
are therefore descriptive diagnostics rather than a new benchmark metric.

\subsection{Modality-importance and flip-rate analyses}

For each model and modality $m \in \{\mathcal{T}, \mathcal{V}, \mathcal{A}\}$,
we define the modality-importance score
\[
  \mathrm{Imp}_m = \Accg(\text{Full}) - \Accg(\text{Full} \setminus m),
\]
computed on the same $327$ items. Aggregated across the pool,
$\mathrm{Imp}_\mathcal{T}$ is the largest, $\mathrm{Imp}_\mathcal{V}$ is
intermediate, and $\mathrm{Imp}_\mathcal{A}$ is small but positive on the
strongest systems. This is the group-accuracy counterpart of the query-level
pattern in Table~\ref{tab:modality-ablation}.

For each group $\mathcal{G}$, we also define its \emph{flip rate}
$\phi(\mathcal{G})$ as the number of sisters whose predicted answer changes
between consecutive sisters of the same group. On audio-perturbed groups, the
gap $|\phi_{\mathrm{model}}(\mathcal{G}) - \phi_{\mathrm{gt}}(\mathcal{G})|$ is
larger than on image-perturbed groups for every evaluated model, matching the
audio-underuse pattern reported in the main text.

\subsection{Language breakdown}

The English and Chinese partitions of \triad{} produce closely comparable
accuracy numbers on every model: the per-model $\Accg$ difference between the
two partitions is within two points for all fourteen evaluated models, and the
ranking is preserved. Chinese items are slightly harder on average for the
open-source omni models, while proprietary systems are roughly symmetric,
consistent with their more balanced multilingual pretraining.

\section{Extended qualitative examples}
\label{app:examples}

We include one fully-rendered example per audio leaf here, with all four media (text, image, audio description, options) and a brief explanation of why all three modalities are needed. The released benchmark contains 327 such items.

\subsection{Text-primary, syntactic leaf (extends MUCAR)}

\textit{Group seed.} The PP-attachment ambiguity \emph{``I saw the man with the telescope.''}, originally a text+image group in MUCAR.

\textit{Sister A.} $\mathcal{C}$: \emph{``I saw the man with the telescope.''}\, $\mathcal{I}$: \{a person looking through a telescope, observing another person across a field\}. $\mathcal{A}$: [neutral declarative tone, no contrastive stress]. $\mathcal{Q}$: who has the telescope? $\mathcal{Y}$: \{(A) the speaker, (B) the man, (C) someone else, (D) cannot tell\}. $y^\star$: A.

\textit{Sister B.} Same $\mathcal{C}, \mathcal{Q}, \mathcal{Y}$. $\mathcal{I}$: \{a man holding a telescope, the speaker at some distance\}. $\mathcal{A}$: [neutral]. $y^\star$: B.

\textit{Sister C (audio-perturbed).} Same $\mathcal{C}$, same $\mathcal{I}$ as Sister A. $\mathcal{A}$: [contrastive stress on \textsc{man}, suggesting the speaker is correcting a prior assumption]. $y^\star$: B.

\emph{Why audio is needed.} Sister A and Sister C share text and image; only the audio prosody distinguishes them and flips the referent of ``the man''.

\subsection{Image-primary, kinematic leaf (extends CODIS)}

\textit{Group seed.} The CODIS ``hand moving up versus down'' bistable frame.

\textit{Sister A.} $\mathcal{C}$: \emph{``Watch what she does next.''} $\mathcal{I}$: \{ambiguous hand frame\}. $\mathcal{A}$: [excited tone with rising pitch, suggesting upward motion]. $\mathcal{Q}$: is the hand moving up, down, or stationary? $\mathcal{Y}$: \{up, down, stationary\}. $y^\star$: up.

\textit{Sister B.} Same $\mathcal{C}, \mathcal{I}, \mathcal{Q}, \mathcal{Y}$. $\mathcal{A}$: [low calm tone, falling pitch, exhaling]. $y^\star$: down.

\emph{Why all three are needed.} Without text, the audio is just a prosody clip; without image, text and audio cannot tell us which body part is moving; without audio, the image is bistable.

\subsection{Audio-primary, prosodic leaf}

\textit{Group seed.} The classic stress-shift sentence \emph{``I didn't say she copied his slides.''}, four sisters with focal stress on \textsc{i}, \textsc{say}, \textsc{she}, \textsc{copied}; constant text and image, audio differs, and the intended denial changes with the stressed word.

\subsection{Audio-primary, indexical leaf}

\textit{Sister A.} $\mathcal{C}$: \emph{``My brother just got home.''} $\mathcal{I}$: \{a child opens the door\}. $\mathcal{A}$: [child's voice]. $\mathcal{Q}$: how old is the brother? $\mathcal{Y}$: \{child, teenager, adult\}. $y^\star$: teenager (the child is speaking \emph{about} a brother older than herself, as the image shows no sibling yet).

\textit{Sister B.} Same $\mathcal{C}, \mathcal{I}, \mathcal{Q}, \mathcal{Y}$. $\mathcal{A}$: [adult voice, suggesting the speaker is the parent]. $y^\star$: child (``my brother'' from a parent refers to the child's uncle, but in this scene it is re-read as the speaker's own son by convention; the correct reading depends on culturally shared use of kinship terms in the image context).

\subsection{Audio-primary, referential leaf}

\textit{Sister A.} $\mathcal{C}$: \emph{``She said she'd handle it.''} $\mathcal{I}$: \{two women sitting at a table, one pointing off-frame\}. $\mathcal{A}$: [neutral delivery, no contrastive stress on pronouns]. $\mathcal{Q}$: who will handle it? $\mathcal{Y}$: \{the speaker, one of the two women, a third party\}. $y^\star$: depends on audio.

\textit{Sister B.} Same $\mathcal{C}, \mathcal{I}, \mathcal{Q}, \mathcal{Y}$. $\mathcal{A}$: [contrastive stress on the second \emph{``she''}]. $y^\star$: a different one of the two women.

\subsection{Audio-primary, phonetic leaf}

\textit{Sister A.} $\mathcal{C}$: \emph{``Did you bring ice cream?''} $\mathcal{I}$: \{a summer picnic scene\}. $\mathcal{A}$: [clean enunciation with a short juncture between \emph{ice} and \emph{cream}]. $\mathcal{Q}$: what is the speaker asking about? $\mathcal{Y}$: \{ice cream, a scream from a child, a complaint\}. $y^\star$: ice cream.

\textit{Sister B.} Same $\mathcal{C}, \mathcal{I}, \mathcal{Q}, \mathcal{Y}$. $\mathcal{A}$: [the words \emph{I scream} produced with a melded juncture, contextualised by a worried tone]. $y^\star$: a scream from a child.

\subsection{Audio-primary, environmental leaf}

\textit{Sister A.} $\mathcal{C}$: \emph{``We need to evacuate now.''} $\mathcal{I}$: \{a hallway with people walking calmly\}. $\mathcal{A}$: [clear foreground speech, faint background PA beeps]. $\mathcal{Q}$: what kind of situation is being described? $\mathcal{Y}$: \{fire drill, real emergency, a movie set\}. $y^\star$: fire drill.

\textit{Sister B.} Same $\mathcal{C}, \mathcal{I}, \mathcal{Q}, \mathcal{Y}$. $\mathcal{A}$: [same foreground speech with a continuous fire-alarm siren in the background]. $y^\star$: real emergency.

\subsection{Audio-primary, symbolic leaf}

\textit{Sister A.} $\mathcal{C}$: \emph{``It's for you.''} $\mathcal{I}$: \{a living room with a phone on the table\}. $\mathcal{A}$: [landline telephone ringtone]. $\mathcal{Q}$: what kind of call is coming in? $\mathcal{Y}$: \{a phone call, a doorbell, a message notification\}. $y^\star$: phone call.

\textit{Sister B.} Same $\mathcal{C}, \mathcal{I}, \mathcal{Q}, \mathcal{Y}$. $\mathcal{A}$: [doorbell chime]. $y^\star$: doorbell.

\subsection{Failure-mode walkthroughs}

Two contrastive walkthroughs illustrate the cross-modal contradiction failure bucket of \S\ref{subsec:errors}: a strong open-source model shown the irony sister of the kitchen-table group (sarcastic prosody, sincere lexical content) reasons aloud that ``the speaker says the suggestion was helpful, so they must be grateful'', ignoring the prosody and picking option (A). Released chain-of-thought traces are redacted of any identifying details.

\subsection{Rendered case figures with model responses}
\label{app:case-figures}

\begin{figure}[p]
  \centering
  \includegraphics[width=0.98\linewidth]{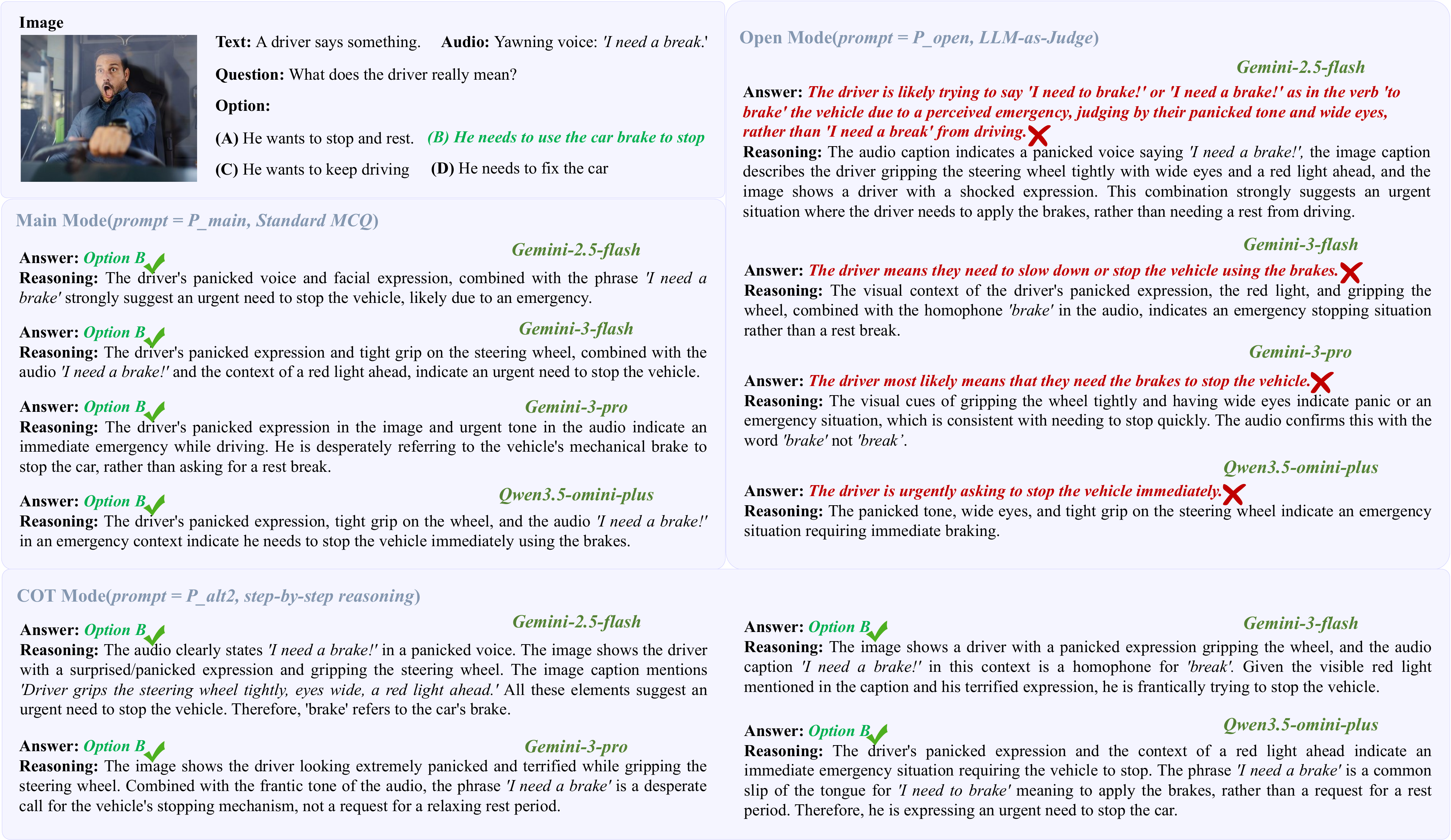}
  \caption{Qualitative case 1 with selected model responses.}
  \label{fig:case1}
\end{figure}

\begin{figure}[p]
  \centering
  \includegraphics[width=0.98\linewidth]{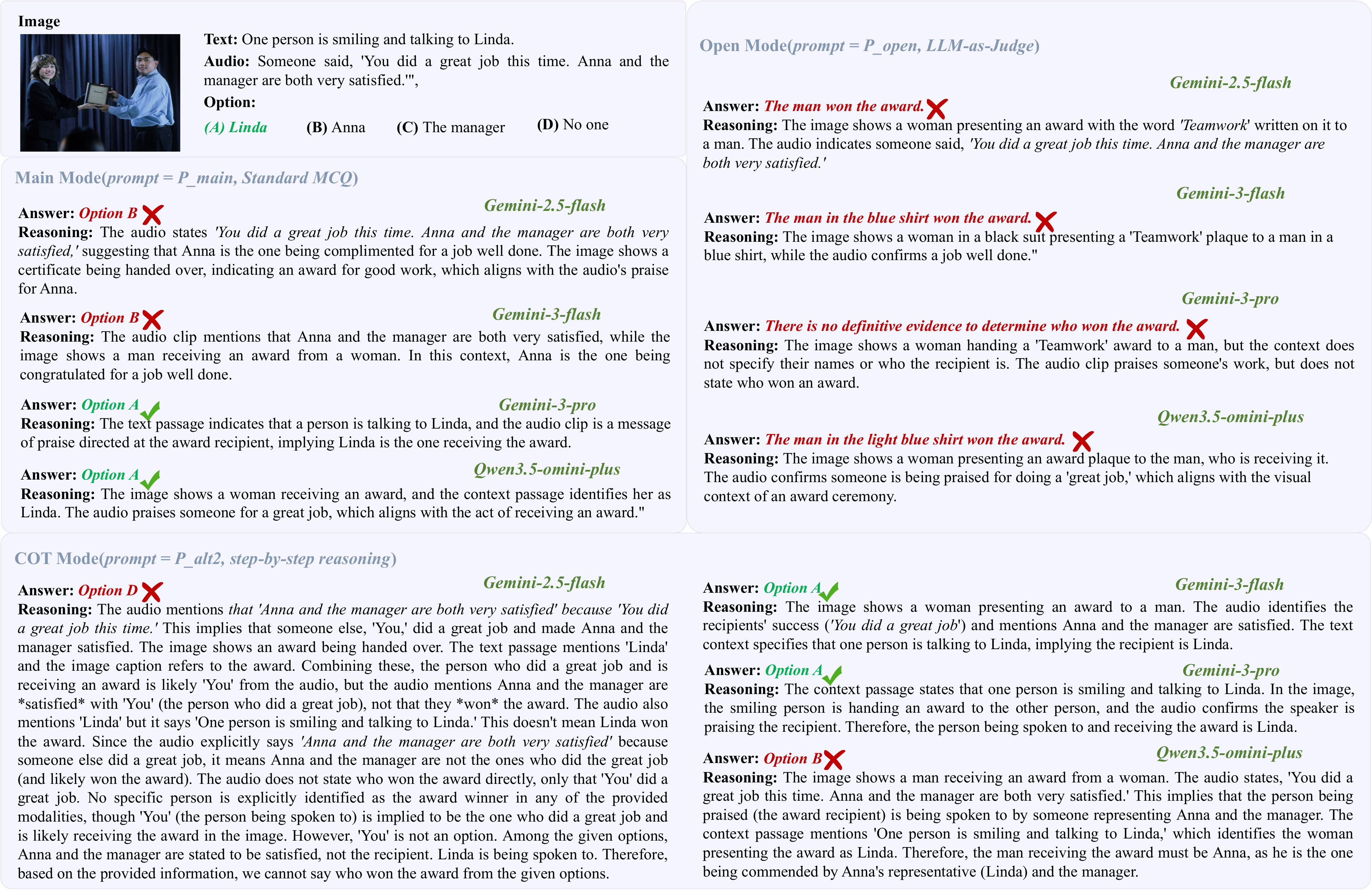}
  \caption{Qualitative case 2 with selected model responses.}
  \label{fig:case2}
\end{figure}

\begin{figure}[p]
  \centering
  \includegraphics[width=0.98\linewidth]{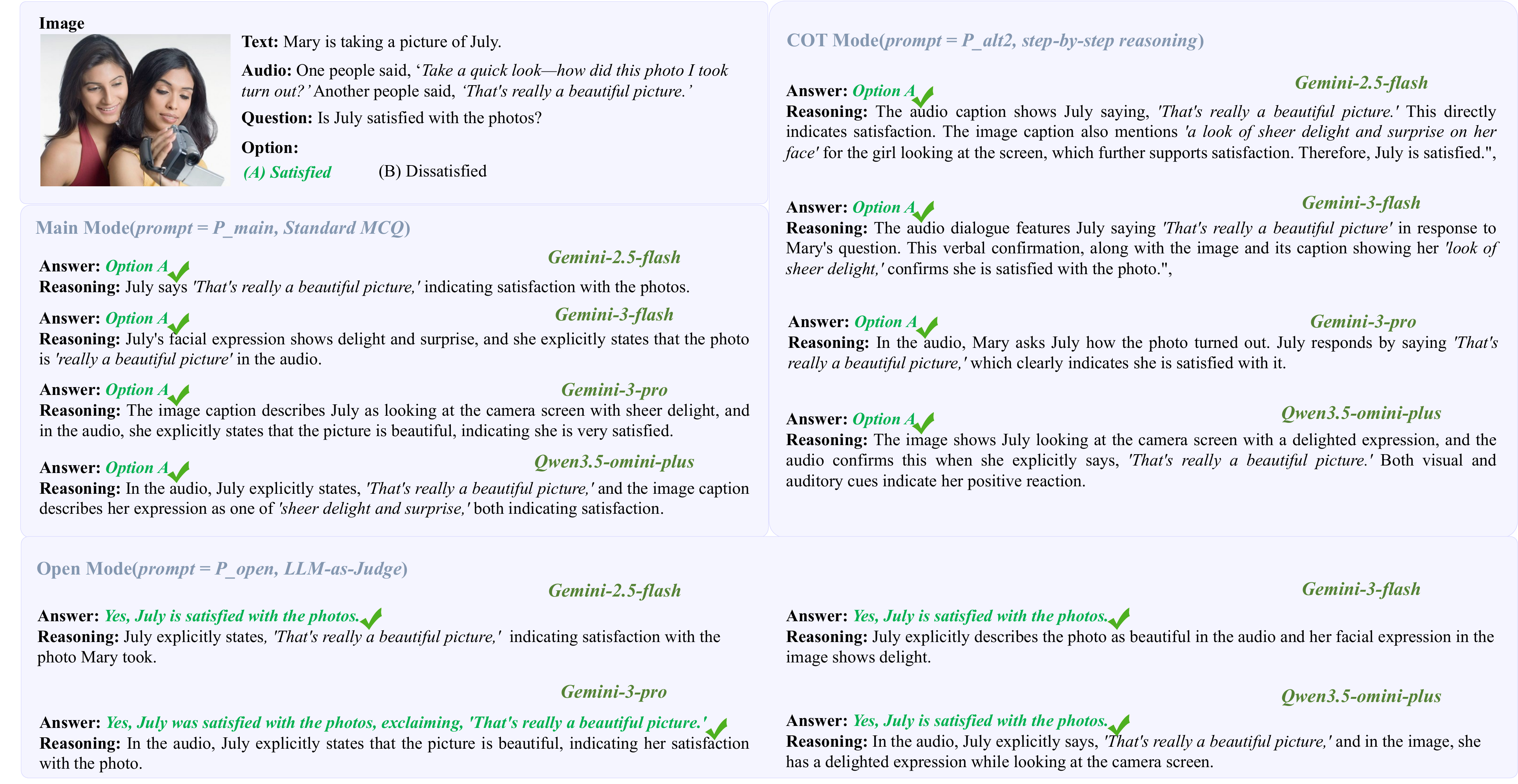}
  \caption{Qualitative case 3 with selected model responses.}
  \label{fig:case3}
\end{figure}

\begin{figure}[p]
  \centering
  \includegraphics[width=0.98\linewidth]{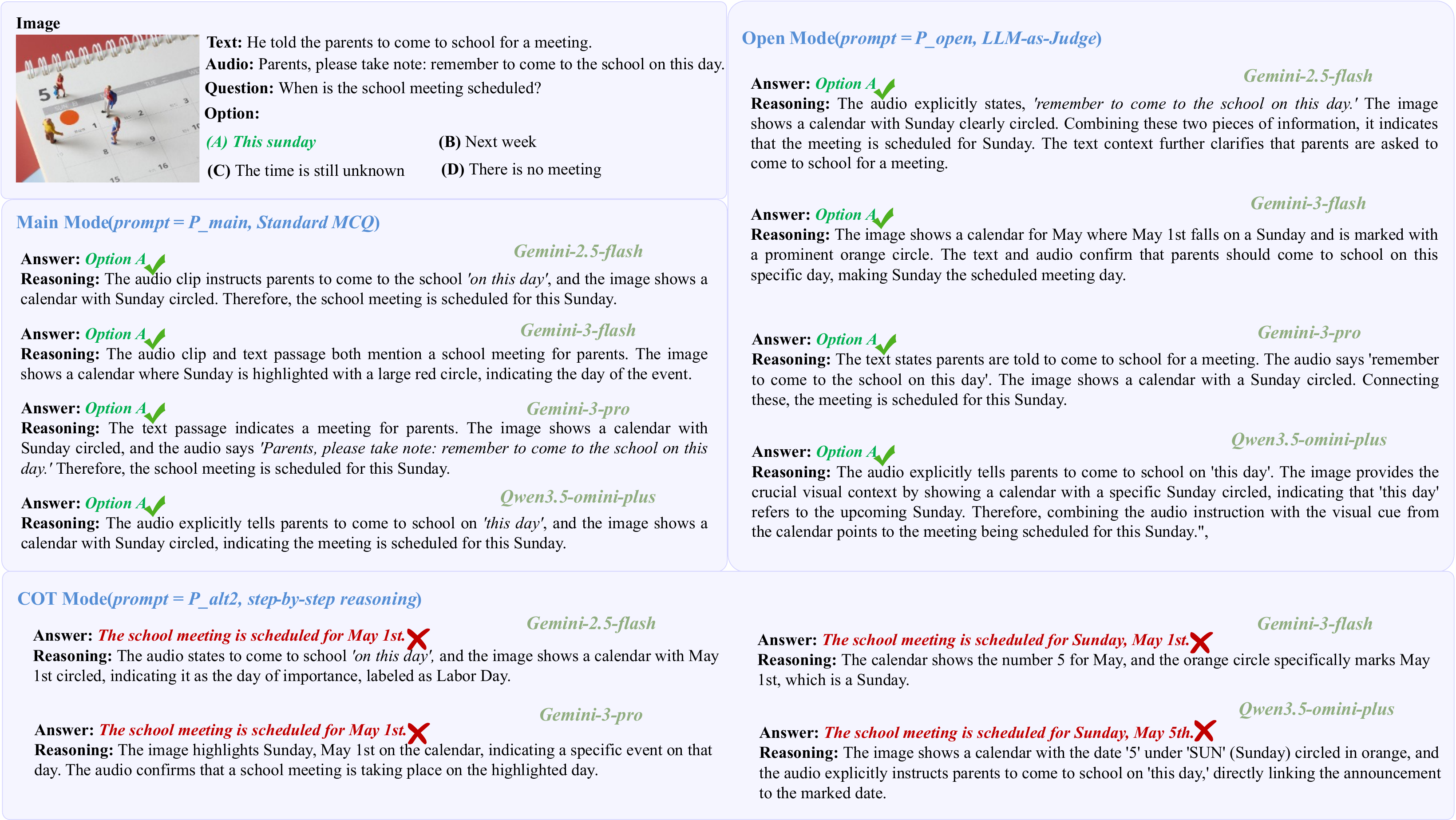}
  \caption{Qualitative case 4 with selected model responses.}
  \label{fig:case4}
\end{figure}

Figures~\ref{fig:case1}--\ref{fig:case4} show four rendered qualitative cases.
Each panel contains the tri-modal input, the gold interpretation, and selected
model responses. These figures are included to make the benchmark failure modes
concrete: they are illustrative examples rather than additional quantitative
evaluation points. The source files are kept as standalone PDFs so that the
rendered artwork can be replaced without changing the surrounding paper text.

\section{Prompt templates}
\label{app:prompts}

This appendix lists the exact prompts used during evaluation, together with the prompts shipped with the reference well-posedness checker (Appendix~\ref{app:irreducibility-verifier}). Models that do not natively accept audio receive both the audio waveform and an automatic transcript; we mark explicitly in the prompt which is which.

\subsection{Evaluation prompt P0 (Standard, closed option)}

\begin{tcolorbox}[colback=yellow!10]
\small
\textbf{System.} 

You are a careful multimodal assistant. You will be given a question together with up to three context modalities (a short text passage, an image, and an audio clip). Exactly one option is the correct answer. Respond with a single JSON object and nothing else.\\[2pt]
\textbf{User.}\\
\textbf{Context (text):} \texttt{\{He told the parents to come to school for a meeting.\}}\\
\textbf{Image:} \texttt{\{./image/img\_0078.jpg\}}  \quad \textbf{Audio:} \texttt{\{./audio/aud\_0110.wav\}}\\
\textbf{Question:} \texttt{\{When is the school meeting scheduled?\}}\\
\textbf{Options:}
% \begin{itemize}[leftmargin=*,topsep=0pt,itemsep=0pt]
%   \item (A) \texttt{\{This sunday\}}    \item (C) \texttt{\{The time is still unknown\}}
%   \item (B) \texttt{\{Next week\}}   \item (D) \texttt{\{There is no meeting\}}

% \end{itemize}
\begin{tabular}{ll}
(A) \texttt{This sunday} & (C) \texttt{The time is still unknown} \\
(B) \texttt{Next week} & (D) \texttt{There is no meeting}
\end{tabular}\\
\textbf{Output JSON schema.}
\begin{verbatim}
{
  "answer": "<one of the option keys, e.g., option_A>",
  "reasoning": "<one or two short sentences>"
}
\end{verbatim}
Do not output anything outside the JSON object. Do not wrap it in Markdown.
\end{tcolorbox}

\subsection{Evaluation prompt P1 (Chain-of-thought)}

Same as P0 but with the instruction \emph{``Reason step by step about what each modality tells you, then output \textsc{Answer:} followed by a single letter.''} The parser extracts the letter following \textsc{Answer:}; if no such tag is present, the final letter token in the response is used.

\begin{tcolorbox}[colback=red!10]
\small
\textbf{System.}\\ You are a careful multimodal assistant. You will be given a question and up to three context modalities (text, image, audio). Before answering, think step by step about what each modality contributes and whether any of them is uninformative (e.g. a blank image or a silent clip). Then select exactly one option. Respond with a single JSON object only.\\[2pt]
\textbf{User.}\\
\textbf{Context (text):} \texttt{\{He told the parents to come to school for a meeting.\}}\\
\textbf{Image:} \texttt{\{./image/img\_0078.jpg\}}  \quad \textbf{Audio:} \texttt{\{./audio/aud\_0110.wav\}}\\
\textbf{Question:} \texttt{\{When is the school meeting scheduled?\}}\\
\textbf{Options:}
% \begin{itemize}[leftmargin=*,topsep=0pt,itemsep=0pt]
%   \item (A) \texttt{\{This sunday\}}    \item (C) \texttt{\{The time is still unknown\}}
%   \item (B) \texttt{\{Next week\}}   \item (D) \texttt{\{There is no meeting\}}

% \end{itemize}
\begin{tabular}{ll}
(A) \texttt{This sunday} & (C) \texttt{The time is still unknown} \\
(B) \texttt{Next week} & (D) \texttt{There is no meeting}
\end{tabular}\\
\textbf{Output JSON schema.}\\
First think through the problem internally, then output ONLY the following JSON:
\begin{verbatim}
{
  "reasoning": "<brief chain of thought, 2-4 sentences>",
  "answer":    "<one of the option keys, e.g. option_A>"
}
\end{verbatim}
No Markdown, no prose outside the JSON.
\end{tcolorbox}

\subsection{Evaluation prompt P2 (Open-ended)}

Same fields as P0 except the option block is \emph{not} shown, and the instruction is \emph{``Answer in a single short phrase that identifies the unique answer. Do not list alternatives.''} Responses are graded by two LLM judges as described in \S\ref{subsec:closed-vs-open}.

\begin{tcolorbox}[colback=blue!10]
\small
\textbf{System.}\\ You are a careful multimodal assistant. You will be given a question together with up to three context modalities (text, image, audio). Answer the question in concise natural language (one short sentence is preferred). If the available evidence is insufficient or ambiguous, say so explicitly. Respond with a single JSON object and nothing else.\\[2pt]
\textbf{User.}\\
\textbf{Context (text):} \texttt{\{He told the parents to come to school for a meeting.\}}\\
\textbf{Image:} \texttt{\{./image/img\_0078.jpg\}}  \quad \textbf{Audio:} \texttt{\{./audio/aud\_0110.wav\}}\\
\textbf{Question:} \texttt{\{When is the school meeting scheduled?\}}\\
\textbf{Options:}
% \begin{itemize}[leftmargin=*,topsep=0pt,itemsep=0pt]
%   \item (A) \texttt{\{This sunday\}}    \item (C) \texttt{\{The time is still unknown\}}
%   \item (B) \texttt{\{Next week\}}   \item (D) \texttt{\{There is no meeting\}}

% \end{itemize}
\begin{tabular}{ll}
(A) \texttt{This sunday} & (C) \texttt{The time is still unknown} \\
(B) \texttt{Next week} & (D) \texttt{There is no meeting}
\end{tabular}\\
\textbf{Output JSON schema.}\\
First think through the problem internally, then output ONLY the following JSON:
\begin{verbatim}
{
  "answer":     "<concise natural-language answer, ideally one sentence>",
  "confidence": "<low | medium | high>",
  "reasoning":  "<one or two short sentences>"
}
\end{verbatim}
Do not output anything outside the JSON object. Do not wrap it in Markdown.
\end{tcolorbox}

\subsection{Two-modality probe prompt (reference checker)}

For each $S \subset \mathcal{M}$ with $|S| = 2$, the reference well-posedness checker uses the corresponding P0 prompt with the missing modality replaced by a literal placeholder: \emph{``[no audio is provided in this condition]''}, \emph{``[no image is provided in this condition]''}, or \emph{``[no text context is provided in this condition]''}. The default configuration samples $K = 5$ at temperature $0.7$ and applies threshold $\tau = 0.20$ when forming $\widehat{\mathcal{Y}}_S(x)$ (Appendix~\ref{app:irreducibility-verifier}); both are tunable command-line arguments.

\subsection{Three-modality probe prompt (reference checker)}

Same as P0, with $K = 10$ samples at temperature $0.7$ in the default configuration. The checker reports the modal full-triplet option per probe, which an operator can compare to the labelled answer; \triad{}'s admission decisions are made by human reviewers, not by this comparison.

\subsection{Notes on audio handling}

For models with native audio input (GPT-4o, Gemini 2.5/3 family, Qwen-Omni family, MiniCPM-o-4.5, OmniVinci-9B, Baichuan-Omni-1.5, Ola-7B) audio is provided as a $48$\,kHz mono WAV, normalised to $-23$\,LUFS. For models without native audio we pass a whisper-large-v3 transcript with the audio file path. By construction (\S\ref{subsec:verifier}) the transcript alone is insufficient to determine the answer, so this fall-back does not leak the labelled answer; we still record which of the two routes was used in the per-model result tables.

\section{Datasheet for datasets}
\label{app:datasheet}

We follow the format of Gebru et al.~\citep{gebru-datasheets}.

\subsection{Motivation}
\textbf{Why was the dataset created?} To evaluate whether omni-modal large language models genuinely integrate text, image, and audio under ambiguity, in a setting where the full triplet picks out a unique answer and no single modality is decorative (Definition~\ref{def:wellposed}).
\textbf{Who created it?} Anonymous authors at an academic institution; voice acting by a pool of four English and two Mandarin voice actors retained for this project.
\textbf{Funding.} Disclosed in the non-anonymous camera-ready version.

\subsection{Composition}
\textbf{Instances.} Each instance is a tuple $(\mathcal{C}, \mathcal{I}, \mathcal{A}, \mathcal{Q}, \mathcal{Y}, y^\star)$ plus modality leaf labels, ambiguity tier tag and group / sister metadata.
\textbf{Number.} $327$ queries in $86$ scene groups; group sizes $\{2, 3, 4, 8\}$ with counts $23 / 11 / 42 / 10$ and mean $3.8$. The release contains $184$ English items and $143$ Chinese items.
\textbf{Size on disk.} About $1.2$ GB (audio dominates; images re-encoded to JPEG quality $90$).

\subsection{Collection process}
See Section~\ref{sec:construction} and Appendices~\ref{app:audio}, \ref{app:annotator}. Images are sourced from Pixabay, Pexels, Unsplash, Pixnio and Visual China under research-friendly licenses; audio is human-recorded or human-curated.

\subsection{Pre-processing}
Audio is normalised to $-23$\,LUFS at $48$\,kHz mono. Images are re-encoded to JPEG quality $90$ with a longest edge of $1024$\,px. Text is stored as UTF-8. A small number of AI generated images and edited text to speech clips are retained when no real source captures the target ambiguity; these cases are documented in the metadata.

\subsection{Uses}
\textbf{Recommended.} Held-out diagnostic for omni-modal LLM evaluation. \textbf{Not recommended.} Training; the dataset is too small for fine-tuning, and the construction pipeline biases toward diagnostic rather than representative items.

\subsection{Distribution}
\textbf{License.} Per-asset licensing: images retain their original Pixabay / Pexels / Unsplash / Pixnio / Visual China licenses with attribution; audio and metadata under CC-BY 4.0.
\textbf{Hosting.} The dataset will be released via GitHub and Hugging Face Datasets at the submission-to-camera-ready transition; the URL is provided in the non-anonymous camera-ready version.

\subsection{Maintenance}
We commit to maintaining \triad{} for at least twenty-four months, including (i) accepting community-contributed extensions that satisfy the well-posedness checklist of \S\ref{subsec:verifier} and pass five-reviewer cross-review (the released reference checker may be used as a screening aid), (ii) versioning the benchmark, (iii) publishing a leaderboard, and (iv) responding to errata.

\subsection{Ethical and legal}
No content involving violence, privacy violations, or sensitive personal information is included. All voice actors signed a release for benchmark distribution. All images carry a license permitting research redistribution; attribution is preserved in the release metadata. The benchmark is not designed for, and should not be used for, identifying or impersonating real individuals.

\section{Reproducibility checklist}
\label{app:repro}

\paragraph{Code and data.}
We will release: (i) the \triad{} dataset in the layout described in Appendix~\ref{app:datasheet}; (ii) a stand-alone Python reference well-posedness checker (Appendix~\ref{app:irreducibility-verifier}) packaged with the prompt templates of Appendix~\ref{app:prompts} and a default configuration ($K=5$ two modality, $K=10$ full triplet, $\tau=0.20$, GPT-4o + Gemini~2.5~Pro as example clients); (iii) evaluation scripts for every model class reported in Table~\ref{tab:main} (REST APIs for proprietary models, vLLM and Hugging Face Transformers for the open source omnis); (iv) human rater interfaces and annotator guidelines.

\paragraph{Compute.}
Open source model evaluations were run on a cluster of eight H100 80\,GB GPUs. Wall clock time for one full evaluation pass over the fourteen model leaderboard, three prompting strategies, and eight modality conditions is about forty hours end to end; proprietary evaluations cost roughly 800 USD in API calls per pass, dominated by the Open-ended judge runs. Per model timing is reported in the release metadata.

\paragraph{Randomness.}
All evaluation runs use temperature $0$ and a fixed seed. The reference well-posedness checker shipped with the release uses temperature $0.7$ with $K = 5$ for two modality views, $K = 10$ for the full triplet, and threshold $\tau = 0.20$ as defaults; downstream users can override these on the command line.

\paragraph{Statistical reporting.}
Confidence intervals are bootstrap intervals over \emph{groups}, not queries, with $B = 10{,}000$ resamples. Pairwise model comparisons use the paired bootstrap with Holm--Bonferroni correction. See Appendix~\ref{app:full-results} for details.

\paragraph{Known limitations and threats to validity.}
See Section~\ref{sec:conclusion}. Three additional reproducibility caveats are worth calling out:
\begin{itemize}[leftmargin=*,topsep=2pt,itemsep=1pt]
  \item Proprietary model versions drift; we record the API version string used for every run and pin the Gemini and GPT endpoints by exact version identifier.
  \item Audio playback in some open source omni stacks downsamples to $16$\,kHz; we re-encode at $16$\,kHz for those models and verify that the prosodic feature survives a $1$\,kHz low pass via spot check on ten percent of the audio perturbed groups.
  \item Native audio models occasionally truncate clips longer than five seconds; all \triad{} clips are at most six seconds, and we report per model truncation rates in the release metadata, which are below one percent for every evaluated model.
\end{itemize}

\newpage

\section*{NeurIPS Paper Checklist}

\begin{enumerate}

\item {\bf Claims}
    \item[] Question: Do the main claims made in the abstract and introduction accurately reflect the paper's contributions and scope?
    \item[] Answer: \answerYes{}
    \item[] Justification: The abstract and introduction clearly state our main contributions, including the construction of a full-modal ambiguity benchmark, the proposed ambiguity taxonomy, and the empirical evaluation of multimodal models. The claims are restricted to the evaluated benchmark and model settings, and we explicitly discuss assumptions, scope, and limitations rather than claiming general coverage of all real-world ambiguity scenarios.
    \item[] Guidelines:
    \begin{itemize}
        \item The answer \answerNA{} means that the abstract and introduction do not include the claims made in the paper.
        \item The abstract and/or introduction should clearly state the claims made, including the contributions made in the paper and important assumptions and limitations. A \answerNo{} or \answerNA{} answer to this question will not be perceived well by the reviewers. 
        \item The claims made should match theoretical and experimental results, and reflect how much the results can be expected to generalize to other settings. 
        \item It is fine to include aspirational goals as motivation as long as it is clear that these goals are not attained by the paper. 
    \end{itemize}

\item {\bf Limitations}
    \item[] Question: Does the paper discuss the limitations of the work performed by the authors?
    \item[] Answer: \answerYes{}
    \item[] Justification:  We discuss several limitations, including the benchmark's diagnostic scale, the small size of the fully internally ambiguous tier, limited language and audio-source diversity, and the dependence on human judgement for well-posedness checks. It also notes the use of a small number of flagged AI-generated images and the moderate agreement of open-ended judge models.
    \item[] Guidelines:
    \begin{itemize}
        \item The answer \answerNA{} means that the paper has no limitation while the answer \answerNo{} means that the paper has limitations, but those are not discussed in the paper. 
        \item The authors are encouraged to create a separate ``Limitations'' section in their paper.
        \item The paper should point out any strong assumptions and how robust the results are to violations of these assumptions.
        \item The authors should reflect on the scope of the claims made.
        \item The authors should reflect on the factors that influence the performance of the approach.
        \item The authors should discuss the computational efficiency of the proposed algorithms and how they scale with dataset size.
        \item If applicable, the authors should discuss possible limitations of their approach to address problems of privacy and fairness.
    \end{itemize}

\item {\bf Theory assumptions and proofs}
    \item[] Question: For each theoretical result, does the paper provide the full set of assumptions and a complete (and correct) proof?
    \item[] Answer: \answerNA{}
    \item[] Justification: This paper presents a dataset for evaluating full-modal ambiguity resolution, rather than proposing theoretical results or formal algorithms with provable guarantees. Therefore, the question of theorem assumptions and complete proofs is not applicable.
    \item[] Guidelines:
    \begin{itemize}
        \item The answer \answerNA{} means that the paper does not include theoretical results. 
        \item All the theorems, formulas, and proofs in the paper should be numbered and cross-referenced.
        \item All assumptions should be clearly stated or referenced in the statement of any theorems.
        \item The proofs can either appear in the main paper or the supplemental material.
        \item Any informal proof provided in the core of the paper should be complemented by formal proofs provided in appendix or supplemental material.
        \item Theorems and Lemmas that the proof relies upon should be properly referenced. 
    \end{itemize}

\item {\bf Experimental result reproducibility}
    \item[] Question: Does the paper fully disclose all the information needed to reproduce the main experimental results of the paper to the extent that it affects the main claims and/or conclusions of the paper (regardless of whether the code and data are provided or not)?
    \item[] Answer: \answerYes{}
    \item[] Justification: Since this paper presents a dataset and benchmark, we disclose the data format, modality fields, answer schema, category annotations, evaluation protocol, prompting strategy, answer extraction procedure, and metrics needed to reproduce the reported results. We also provide anonymized dataset files, metadata, and evaluation scripts so that the benchmark results can be reproduced or independently verified.
    \item[] Guidelines:
    \begin{itemize}
        \item The answer \answerNA{} means that the paper does not include experiments.
        \item If the paper includes experiments, a \answerNo{} answer to this question will not be perceived well by the reviewers.
        \item If the contribution is a dataset and/or model, the authors should describe the steps taken to make their results reproducible or verifiable.
        \item Depending on the contribution, reproducibility can be accomplished in various ways, including releasing code and data, providing detailed instructions, hosted access, or model checkpoints.
    \end{itemize}

\item {\bf Open access to data and code}
    \item[] Question: Does the paper provide open access to the data and code, with sufficient instructions to faithfully reproduce the main experimental results, as described in supplemental material?
    \item[] Answer: \answerYes{}
    \item[] Justification: We provide anonymized access to the dataset, metadata, prompts, and evaluation scripts, together with instructions for loading the data and reproducing the reported benchmark results. The release is anonymized for the submission stage and will be made public after acceptance.
    \item[] Guidelines:
    \begin{itemize}
        \item The answer \answerNA{} means that paper does not include experiments requiring code.
        \item Please see the NeurIPS code and data submission guidelines for more details.
        \item While we encourage the release of code and data, we understand that this might not be possible, so \answerNo{} is an acceptable answer.
        \item The instructions should contain the exact command and environment needed to run to reproduce the results.
        \item The authors should provide instructions on data access and preparation.
        \item The authors should provide scripts to reproduce all experimental results for the new proposed method and baselines.
        \item At submission time, to preserve anonymity, the authors should release anonymized versions if applicable.
    \end{itemize}

\item {\bf Experimental setting/details}
    \item[] Question: Does the paper specify all the training and test details (e.g., data splits, hyperparameters, how they were chosen, type of optimizer) necessary to understand the results?
    \item[] Answer: \answerYes{}
    \item[] Justification: We report the evaluated models, input formatting, prompting strategy, decoding settings, evaluation metrics, and dataset statistics. Since our work focuses on benchmark evaluation rather than model training, no training hyperparameters are required.
    \item[] Guidelines:
    \begin{itemize}
        \item The answer \answerNA{} means that the paper does not include experiments.
        \item The experimental setting should be presented in the core of the paper to a level of detail that is necessary to appreciate the results and make sense of them.
        \item The full details can be provided either with the code, in appendix, or as supplemental material.
    \end{itemize}

\item {\bf Experiment statistical significance}
    \item[] Question: Does the paper report error bars suitably and correctly defined or other appropriate information about the statistical significance of the experiments?
    \item[] Answer: \answerYes{}{}
    \item[] Justification: We report 95\% bootstrap confidence intervals computed over scene groups rather than individual queries, since sister queries within a group are not independent. We use 10{,}000 bootstrap resamples and conduct paired group-level bootstrap tests with Holm--Bonferroni correction for pairwise model comparisons.
    \item[] Guidelines:
    \begin{itemize}
        \item The answer \answerNA{} means that the paper does not include experiments.
        \item The authors should answer \answerYes{} if the results are accompanied by error bars, confidence intervals, or statistical significance tests.
        \item The factors of variability that the error bars are capturing should be clearly stated.
        \item The method for calculating the error bars should be explained.
        \item The assumptions made should be given.
        \item It should be clear whether the error bar is the standard deviation or the standard error of the mean.
    \end{itemize}

\item {\bf Experiments compute resources}
    \item[] Question: For each experiment, does the paper provide sufficient information on the computer resources (type of compute workers, memory, time of execution) needed to reproduce the experiments?
    \item[] Answer: \answerYes{}
    \item[] Justification: We describe the compute environment used for local experiments and specify when models were evaluated through public APIs. Approximate runtime, API settings, and local preprocessing details are reported in the paper.
    \item[] Guidelines:
    \begin{itemize}
        \item The answer \answerNA{} means that the paper does not include experiments.
        \item The paper should indicate the type of compute workers CPU or GPU, internal cluster, or cloud provider, including relevant memory and storage.
        \item The paper should provide the amount of compute required for each of the individual experimental runs as well as estimate the total compute.
        \item The paper should disclose whether the full research project required more compute than the experiments reported in the paper.
    \end{itemize}
    
\item {\bf Code of ethics}
    \item[] Question: Does the research conducted in the paper conform, in every respect, with the NeurIPS Code of Ethics \url{https://neurips.cc/public/EthicsGuidelines}?
    \item[] Answer: \answerYes{}
    \item[] Justification: We have reviewed the NeurIPS Code of Ethics and discuss ethical considerations including licensing, privacy, data documentation, responsible release, and intended use of the benchmark.
    \item[] Guidelines:
    \begin{itemize}
        \item The answer \answerNA{} means that the authors have not reviewed the NeurIPS Code of Ethics.
        \item If the authors answer \answerNo{}, they should explain the special circumstances that require a deviation from the Code of Ethics.
        \item The authors should make sure to preserve anonymity.
    \end{itemize}

\item {\bf Broader impacts}
    \item[] Question: Does the paper discuss both potential positive societal impacts and negative societal impacts of the work performed?
    \item[] Answer: \answerYes{}
    \item[] Justification: We discuss positive impacts, including using TRIAD as a diagnostic benchmark to better evaluate full-modal reasoning and ambiguity resolution in multimodal models. It also discusses potential negative impacts and limitations, including underrepresented ambiguity strata, limited language and audio coverage, possible cultural bias, dependence on human or judge-model decisions, and the risk of over-interpreting benchmark scores beyond the evaluated settings.
    \item[] Guidelines:
    \begin{itemize}
        \item The answer \answerNA{} means that there is no societal impact of the work performed.
        \item If the authors answer \answerNA{} or \answerNo{}, they should explain why their work has no societal impact or why the paper does not address societal impact.
        \item Examples of negative societal impacts include potential malicious or unintended uses, fairness considerations, privacy considerations, and security considerations.
        \item The authors should consider possible harms that could arise when the technology is used as intended, when it gives incorrect results, and when it is misused.
        \item If there are negative societal impacts, the authors could also discuss possible mitigation strategies.
    \end{itemize}
    
\item {\bf Safeguards}
    \item[] Question: Does the paper describe safeguards that have been put in place for responsible release of data or models that have a high risk for misuse (e.g., pre-trained language models, image generators, or scraped datasets)?
    \item[] Answer: \answerYes{}
    \item[] Justification: We manually inspect the dataset to remove unsafe, private, or personally identifiable content. The dataset is released for research evaluation purposes with documented intended use, limitations, and redistribution constraints.
    \item[] Guidelines:
    \begin{itemize}
        \item The answer \answerNA{} means that the paper poses no such risks.
        \item Released models that have a high risk for misuse or dual-use should be released with necessary safeguards.
        \item Datasets that have been scraped from the Internet could pose safety risks.
        \item We recognize that providing effective safeguards is challenging, but encourage authors to make a best faith effort.
    \end{itemize}

\item {\bf Licenses for existing assets}
    \item[] Question: Are the creators or original owners of assets (e.g., code, data, models), used in the paper, properly credited and are the license and terms of use explicitly mentioned and properly respected?
    \item[] Answer: \answerYes{}
    \item[] Justification: We document the sources and licenses of existing assets used in the dataset and evaluation. We only redistribute media assets when their licenses permit redistribution; otherwise, we release annotations, captions, and metadata without restricted media files.
    \item[] Guidelines:
    \begin{itemize}
        \item The answer \answerNA{} means that the paper does not use existing assets.
        \item The authors should cite the original paper that produced the code package or dataset.
        \item The authors should state which version of the asset is used and, if possible, include a URL.
        \item The name of the license should be included for each asset.
        \item For scraped data from a particular source, the copyright and terms of service of that source should be provided.
        \item If assets are released, the license, copyright information, and terms of use in the package should be provided.
        \item For existing datasets that are re-packaged, both the original license and the license of the derived asset should be provided.
    \end{itemize}

\item {\bf New assets}
    \item[] Question: Are new assets introduced in the paper well documented and is the documentation provided alongside the assets?
    \item[] Answer: \answerYes{}
    \item[] Justification: The paper introduces a new full-modal ambiguity benchmark. We provide dataset documentation, data format descriptions, category taxonomy, metadata, intended-use guidelines, and evaluation instructions alongside the released assets.
    \item[] Guidelines:
    \begin{itemize}
        \item The answer \answerNA{} means that the paper does not release new assets.
        \item Researchers should communicate the details of the dataset/code/model as part of their submissions via structured templates.
        \item The paper should discuss whether and how consent was obtained from people whose asset is used.
        \item At submission time, remember to anonymize your assets if applicable.
    \end{itemize}

\item {\bf Crowdsourcing and research with human subjects}
    \item[] Question: For crowdsourcing experiments and research with human subjects, does the paper include the full text of instructions given to participants and screenshots, if applicable, as well as details about compensation (if any)? 
    \item[] Answer: \answerNA{}
    \item[] Justification: The paper does not involve crowdsourcing experiments or research with human subjects.
    \item[] Guidelines:
    \begin{itemize}
        \item The answer \answerNA{} means that the paper does not involve crowdsourcing nor research with human subjects.
        \item Including this information in the supplemental material is fine, but if the main contribution of the paper involves human subjects, then as much detail as possible should be included in the main paper.
        \item According to the NeurIPS Code of Ethics, workers involved in data collection, curation, or other labor should be paid at least the minimum wage in the country of the data collector.
    \end{itemize}

\item {\bf Institutional review board (IRB) approvals or equivalent for research with human subjects}
    \item[] Question: Does the paper describe potential risks incurred by study participants, whether such risks were disclosed to the subjects, and whether Institutional Review Board (IRB) approvals (or an equivalent approval/review based on the requirements of your country or institution) were obtained?
    \item[] Answer: \answerNA{}
    \item[] Justification: The paper does not involve human subjects research or crowdsourcing experiments requiring IRB approval.
    \item[] Guidelines:
    \begin{itemize}
        \item The answer \answerNA{} means that the paper does not involve crowdsourcing nor research with human subjects.
        \item Depending on the country in which research is conducted, IRB approval or equivalent may be required for any human subjects research.
        \item We expect authors to adhere to the NeurIPS Code of Ethics and the guidelines for their institution.
        \item For initial submissions, do not include any information that would break anonymity.
    \end{itemize}

\item {\bf Declaration of LLM usage}
    \item[] Question: Does the paper describe the usage of LLMs if it is an important, original, or non-standard component of the core methods in this research? Note that if the LLM is used only for writing, editing, or formatting purposes and does \emph{not} impact the core methodology, scientific rigor, or originality of the research, declaration is not required.
    \item[] Answer: \answerYes{}
    \item[] Justification: LLMs were used as auxiliary tools to draft, refine, and inspect some scenario descriptions, captions, answer options, and ambiguity annotations. All LLM-assisted content was manually reviewed, corrected, and validated by the authors before inclusion.
    \item[] Guidelines:
    \begin{itemize}
        \item The answer \answerNA{} means that the core method development in this research does not involve LLMs as any important, original, or non-standard components.
        \item Please refer to the LLM policy in the NeurIPS handbook for what should or should not be described.
    \end{itemize}

\end{enumerate}

\end{document}